\documentclass[journal]{IEEEtran}

\usepackage{multirow}

\usepackage{url}
\usepackage{cite}
\usepackage{bigstrut}
\usepackage{amsthm}
\usepackage{graphics} 
\usepackage{graphicx}
\usepackage{epsfig}
\usepackage{amssymb}
\usepackage{amsmath}
\usepackage{colortbl}
\usepackage{subfig}
\usepackage{epstopdf}
\usepackage{graphicx,psfrag}
\usepackage{xcolor}
\usepackage{array}
\usepackage{supertabular}
\usepackage{footnote}
\usepackage{indentfirst}
\usepackage[linesnumbered,titlenumbered,ruled]{algorithm2e}
\usepackage{multirow,booktabs,color,soul,threeparttable} 

\newcommand{\tabincell}[2]{\begin{tabular}{@{}#1@{}}#2\end{tabular}}
\newtheorem{definition}{\textbf{Definition}}
\DeclareMathOperator*{\argmax}{argmax}
\DeclareMathOperator*{\argmin}{argmin}
\SetKwInOut{Input}{Input}
\SetKwInOut{Output}{Output}

\begin{document}
\title{Locating the boundaries of Pareto fronts: A Many-Objective Evolutionary Algorithm Based on Corner Solution Search}
\author{  Xinye~Cai, Haoran~Sun, Chunyang~Zhu, Zhenyu Li, Qingfu Zhang  
\thanks{This paper is submitted for possible publication. Reviewers can use this manuscript as an alternative in peer review.}
\thanks{ Xinye~Cai, Haoran~Sun, Chunyang~Zhu and Zhenyu Li are with the College of Computer Science and Technology, Nanjing University of Aeronautics and Astronautics, Nanjing, Jiangsu, 210016 P. R. China; and also with the Collaborative Innovation Center of Novel Software Technology and Industrialization, Nanjing 210023, China (e-mail: xinye@nuaa.edu.cn, nuaa\_sunhr@yeah.net).}
\thanks{Qingfu Zhang is with the Department of Computer Science, City University of Hong Kong, Kowloon, Hong Kong SAR (e-mail: qingfu.zhang@cityu.edu.hk). }
\thanks{This work was supported in part by the National Natural Science Foundation of China (NSFC) under grant 61300159, by ANR/RGC Joint Research Scheme under grant A-CITYU101/16, by the Natural Science Foundation of Jiangsu Province of China under grant BK20130808 and by China Postdoctoral Science Foundation under grant 2015M571751.}}

\maketitle

\begin{abstract}

In this paper, an evolutionary many-objective optimization algorithm based on corner solution search (MaOEA-CS) was proposed. MaOEA-CS implicitly contains two phases: the exploitative search for the most important boundary optimal solutions -- corner solutions, at the first phase, and the use of angle-based selection~\cite{CaiYFZ16} with the explorative search for the extension of PF approximation at the second phase. Due to its high efficiency and robustness to the shapes of PFs, it has won the  CEC$^\prime$2017 Competition on Evolutionary Many-Objective Optimization. In addition, MaOEA-CS has also been applied on two real-world engineering optimization problems with very irregular PFs. The experimental results show that MaOEA-CS outperforms other six state-of-the-art compared algorithms, which indicates it has the ability to handle real-world complex optimization problems with irregular PFs.
\end{abstract}
\IEEEpeerreviewmaketitle


\section{Introduction}
A multiobjective optimization problem (MOP) can be defined as follows:
\begin{eqnarray}
&\mbox{minimize} & F(x)=(f_{1}(x), \ldots, f_{m}(x))^T \label{MOP}\\
\nonumber & \mbox{subject to} & x \in \Omega
\end{eqnarray}
where $\Omega$ is the \emph{decision space}, $F:$ $\Omega\rightarrow R^{m}$ consists of $m$ real-valued objective functions. The \emph{attainable objective set} is $\{F(x)| x \in \Omega \}$. Let $u, v \in R^{m}$, $u$ is said to \emph{dominate} $v$, denoted by $u \prec v $, if and only if $u_{i} \leq v_{i}$ for every $i \in \{1,\ldots, m\}$ and $u_{j} < v_{j} $ for at least one index $j \in \{1, \ldots, m\}$\footnote{In the case of maximization, the inequality signs should be reversed.}. A solution $x^{\ast} \in \Omega$ is \emph{Pareto-optimal} to (\ref{MOP}) if there exists no solution $x\in \Omega$ such that $F(x)$ dominates $F(x^\ast)$. The set of all the Pareto-optimal points is called the \emph{Pareto set} (\emph{PS}) and the set of all the Pareto-optimal objective vectors is the \emph{Pareto front} (\emph{PF}) \cite{Miettinen99}. MOPs with more than three objectives are commonly referred to as many-objective optimization problems (MaOPs).
%

Over the past decades, multi-objective evolutionary algorithms (MOEAs) have been recognized as a major methodology to approximate PFs in MOPs~\cite{Coello2006Evolutionary, Coello200620, Deb2002A, Zhang07b, Deb2001Multi, Fonseca2014An}. However, most MOEAs are designed to address MOPs with two or three objectives. It is well-known that the performance of MOEAs, especially Pareto-dominance based MOEAs, deteriorate when dealing with MaOPs with more than three objectives.  Generally speaking, MaOPs are very challenging due to the following reasons~\cite{Ishibuchi08Evolutionary,Li15Many}.

\begin{enumerate}
  \item With the increase of the number of objectives, the selection pressure of Pareto dominance-based MOEAs deteriorates rapidly, as most solutions become nondominated to each other~\cite{Fonseca98Multiobjective, Purshouse07On, Knowles07Quantifying, Corne2007Techniques}. For example, Pareto dominance-based MOEAs such as NSGA-II~\cite{Deb2002A} and SPEA2~\cite{Zitzler2001SPEA2} can not perform well on MaOPs.
  \item The PF of an $m$-objective non-degenerate MOP (or MaOP) is an ($m-1$)-dimensional manifold~\cite{Ishibuchi15Behavior, Jin03Connectedness} (PFs of degenerate MOPs (or MaOPs) are less than ($m-1$)-dimensional). This indicates that maintaining diversity with a limited number of solutions for MaOPs become more and more difficult. A good example is that the diversity maintenance method used in MOEAs, such as the crowding distance in NSGA-II~\cite{Deb2002A} has become ineffective for MaOPs.
\end{enumerate}

Over the recent years, a large number of many objective optimization evolutionary algorithms (MaOEAs) have been proposed to address MaOPs~\cite{Bader2011HypE, Li2015An, He2016Many, Wang2015Two, Yang2013A, Zhang2015A, Cheng2016A, Deb2014An}. Based on the selection of solutions, they can be roughly divided to following four categories.

\begin{enumerate}
  \item Modified-Pareto-dominance-based approaches directly modify the Pareto-dominance relation to further enhance the selection pressure towards PFs for MaOPs. $\epsilon$-dominance~\cite{Laumanns2002Combining, Hadka2013Borg}, grid-dominance~\cite{Yang13}, volume-dominance~\cite{Le2009An}, and subspace-dominance~\cite{Hern2013Adaptive, Aguirre2010A} belong to this type of approaches.
  \item Diversity-based approaches further enhance the selection pressure of MOEAs by maintaining better diversity. For instance,a diversity management mechanism based on the spread of the population was introduced in~\cite{Adra11}. A shift-based density estimation (SDE) was proposed as a diversity maintenance scheme to further enhance the selection pressure in~\cite{Li14} .
  \item Indicator-based approaches adopt the indicator metric as the selection criteria. For example, hypervolume~\cite{Zitzler99c} is a well-known indicator, which considers both convergence and diversity. In~\cite{Emmerich05Single}, a $S$-metric selection by maximizing the hypervolume of the solution sets was proposed. To further reduce the computational complexity of calculating hypervolume, Bader et al. proposed a hypervolume estimation algorithm (HypE)~\cite{Bader2011HypE}, where, instead of calculating the exact values of hypervolume, Monte Carlo sampling is adopted to approximate it.
  \item Decomposition-based approaches decomposes a multiobjective optimization problem into a number of subproblems by linear or nonlinear aggregation functions and solve them in a collaborative manner. Multiobjective evolutionary algorithm based on decomposition (MOEA/D)~\cite{Zhang07b} is a representative of such approaches. Recent research indicates that decomposition-based approaches~\cite{Ishibuchi09Evolutionary, Ishibuchi15Behavior} (e.g., MOEA/D~\cite{Zhang07b}) has very good performance on MaOPs. However, the diversity of MOEA/D is maintained by a set of preset direction vectors. Very recent research has shown that performance of decomposition-based many-objective algorithms strongly depends on Pareto front shapes~\cite{Ishibuchi2016Performance}. It is difficult for MOEA/D to maintain diversity when the shapes of PFs are irregular~\cite{Cai2017A}.
\end{enumerate}

Some recent works focus on the hybridization of decomposition and dominance approaches~\cite{Deb2014An, Li2015An, Singh2011A, Gong2013Set, He2016Many}. For instance, Deb et al.~\cite{Deb2014An, Jain2014An} proposed a reference-point-based many-objective evolutionary algorithm (NSGA-III) as an extension of NSGA-II~\cite{Deb2002A}. A many-objective evolutionary algorithm based on both dominance and decomposition is also proposed to address MaOPs~\cite{Li2015An}.


More recently, fifteen test problems with different shapes of PFs were proposed for CEC$^\prime$2017 Competition on Evolutionary Many-Objective Optimization in~\cite{cheng2017benchmark}. This test suite aims to promote the research of MaOEAs via suggesting a set of test problems with a good representation of various real-world scenarios.

To address these problems, an evolutionary many-objective optimization algorithm based on corner solution search (MaOEA-CS) was proposed. MaOEA-CS implicitly contains two phases: the exploitative search for the solutions of the most important subproblems (containing corner solutions) at the first phase, and the use of angle-based selection~\cite{CaiYFZ16} with the explorative search for the extension of PF approximation at the second phase. Due to its high efficiency and robustness to the shapes of PFs, it has won the  CEC$^\prime$2017 Competition on Evolutionary Many-Objective Optimization~\footnote{\url{http://www.cercia.ac.uk/news/cec2017maooc/}}.

The rest of this paper is organized as follows. Section II introduces the motivations of MaOEA-CS. Section III elaborates MaOEA-CS. Section IV presents the experimental studies of MaOEA-CS on 15 MaOPs with the different number of objectives for CEC$^\prime$2017 competition~\cite{cheng2017benchmark}. Section V concludes this paper.

\section{Definitions and motivations}
\subsection{Definitions}

An MOP can be decomposed into a number of single objective optimization subproblems to be solved simultaneously in a collaborative way. A representative of such approaches is MOEA/D~\cite{Zhang07b} and its variants~\cite{Cai2015An,Cai2016Decomposition,Cai2017A,Cai2017B}. One of the most commonly used decomposition methods~\cite{Miettinen99} is Weighted Sum.
Let $\lambda=(\lambda_1,\ldots,\lambda_m)^T$ be a direction vector for a subproblem, where $\lambda_i\geq 0$, $i \in \{1, \ldots, m\}$. 

\begin{enumerate}
\item \textbf{Weighted Sum (WS):} A subproblem is defined as

\begin{equation}\label{decomposition1}
    \begin{aligned}
    &\mbox{minimize} && g^{ws}(x|\lambda) = \sum_{i =1}^m{\lambda_i f_i(x)} \; ,\\
    &\mbox{subject to} && x\in \Omega \; .
    \end{aligned}
\end{equation}
\end{enumerate}

To explain our motivations, some notations, such as the boundary direction vector, can be defined as follows.
\begin{definition}[Boundary direction vector]
    If a direction vector $\lambda = (\lambda_1,\dots,\lambda_m)^T$ satisfies:
    \begin{equation}\label{equ:direct-vector}
      \exists i \in \{1,\dots,m\}, \lambda_i = 0;
    \end{equation}
    Then such a vector is called a boundary direction vector.
\end{definition}

The boundary direction vectors are the vectors with zero value for at least one objective. For instance, for a bi-objective optimization problem, the boundary direction vectors are distributed along the coordinate axis. For an $m$-objective optimization problem, the boundary direction vectors are distributed on any (m-1)-dimensional hyperplane.

Based on the boundary direction vectors, we can further define the boundary optimal solutions as follows.
\begin{definition}[Boundary optimal solution]
    Given any a boundary direction vector $\lambda = (\lambda_1,\dots,\lambda_m)^T$, a boundary optimal solution $x=(x_1,\dots,x_m)^T$ can be defined as follows.
    \begin{equation}\label{equ:boundary-solution}
        x = \argmin_{x \in {PS}} g^{ws}(x|\lambda)
    \end{equation}
\end{definition}

Obviously, the boundary optimal solutions are located on the boundaries of a PF, which contain much more information with regard to convergence than other Pareto optimal solutions. However, the number of boundary optimal solutions increase exponentially with the increase of the number of objectives. Under this circumstance, it is more practical to use some representative ones to approximate the PF boundaries. These solutions are called corner solutions, which are obtained by corner direction vectors.

\begin{definition}[Corner direction vectors]
For an $m$-objective optimization problem, the corner direction vectors are two groups of special boundary direction vectors, $V_1$ and $V_2$, as follows.

\begin{equation}\label{V1}
  V_1 = \{\lambda|\exists i \in \{1,2,\dots,m\}, \lambda_i = 0 \wedge \forall j\neq i, \lambda_j = 1.\}
\end{equation}

\begin{equation}\label{V2}
  V_2 = \{\lambda|\exists i \in \{1,2,\dots,m\}, \lambda_i = 1 \wedge \forall j\neq i, \lambda_j = 0.\}
\end{equation}
\end{definition}

Based on Eq.~\ref{V1}, only one element in a direction vector of $V_1$ is 0 and the other ($m-1$) elements are 1. On the contrary, only one element in a direction vector of $V_2$ is 1 and the other ($m-1$) elements are 0, based on Eq.~\ref{V2}.

\begin{definition}[Pareto corner solution]
A set of (Pareto) corner solutions $P$ are the boundary optimal solutions for the corner direction vector $V_1\cup V_2$.
\end{definition}
For an $m$-objective optimization problem, the size of $V_1\cup V_2$ is less than $2m$, thus the size of $P$ is also less than $2m$, where $m$ is the number of objectives. It is worth noting that multiple corner direction vectors may lead to the same corner solution.

\subsection{motivations}
Based on the concept of corner solutions, a two-phase MaOEA-CS is motivated with the following two considerations.
\begin{enumerate}
  \item In the decomposition-based MOEAs, all the subproblems are treated equally important. However, the subproblems containing Pareto corner solutions are apparently more important as the corner solutions can be used to locate the ranges of PFs and help the convergence of other subproblems. Therefore, the exploitative search is favorable to be applied on them for obtaining the important corner solutions in the first phase.
  \item After the Pareto corner solutions are approximated, the explorative search can be conducted for the extension of PF approximation in the second phase. Combined with the use of corner solutions for maintaining the convergence, the angle-based selection~\cite{CaiYFZ16}, which is robust to the shapes of PFs, is adopted for the further diversity improvement.
\end{enumerate}

\section{MaOEA-CS}
Based on the motivations in the last section, MaOEA-CS is proposed and elaborated in this section.

\subsection{Corner solution search}

In the first phase of MaOEA-CS, two sets of the corner direction vectors are used to approximate two sets of corner solutions $P_1$ and $P_2$. Each solution in $P_1$ are closest to one of the $m$ coordinate axis as follows.
\begin{equation}\label{equ:axis}
  P_1=\{x|x = \argmin_{x \in P}{dist^{\perp}(F(x),e^i), i=1,2,\dots,m}\}
\end{equation}
where $dist^{\perp}(a,b)$ indicates the the perpendicular distance from a vector $a$ to a direction vector $b$; $e^i$ denotes the direction vector along $i$-th axis; and $P$ is a nondominated set.

The solutions in $P_2$ are closest to $m$ hypersurfaces determined by arbitrary $(m-1)$ coordinate axis, as follows.

\begin{equation}\label{equ:plane}
  P_2=\{x|x = \argmin_{x \in P}{f_i(x),i=1,2,\dots,m}\}
\end{equation}

By combining $P_1$ and $P_2$, the corner solution set $P_c = P_1 \cup P_2$ can be obtained. After the corner solution set $P_c$ ($|P_c| \leq 2m$) is approximated, the nadir point $z^{nad}$ can be further approximated as follows.
\begin{equation}\label{equ:nad}
   \begin{aligned}
    & z^{nad} = (z_1^{nad},z_2^{nad},\dots,z_m^{nad})^T,\\
    & \mbox{where~~~} z_i^{nad} = \max_{x \in PS}{f_i(x)}  \approx  \max_{x \in P_c}{f_i(x)}.
  \end{aligned}
\end{equation}

The whole procedures of the corner solution search is given in Algorithm \ref{alg:cs}. $P_1$ is firstly obtained based on Eq.~(\ref{equ:axis}) and $z^{nad}$ is approximated by $P_1$ based on Eq.~(\ref{equ:nad}). After that, $P_2$ is obtained based on Eq.~(\ref{equ:plane}). If the objective value of any solution $x$ in $P_2$ is larger than that of the approximated $z^{nad}$, $x$ is also added to $P_c$ as a corner solution.

Algorithm \ref{alg:cs} can be regarded as the combination of Pareto dominance and decomposition~\cite{Zhang07b} only using $2m$ important boundary direction vectors. A very natural extension is to consider more boundary direction vectors (i.e., subproblems in MOEA/D~\cite{Zhang07b}) to locate more boundaries of the PFs, which can help approximate nadir point more accurately for MaOPs with very irregular PFs. However, the appropriate balance between more subproblems for better coverage of boundary PFs and the affordable computational cost should be carefully considered, which could be an interesting research direction for the future.

\begin{figure*}
  \centering
  \subfloat[]{
      \label{fig:dtlz2}
      \includegraphics[width=1.8in]{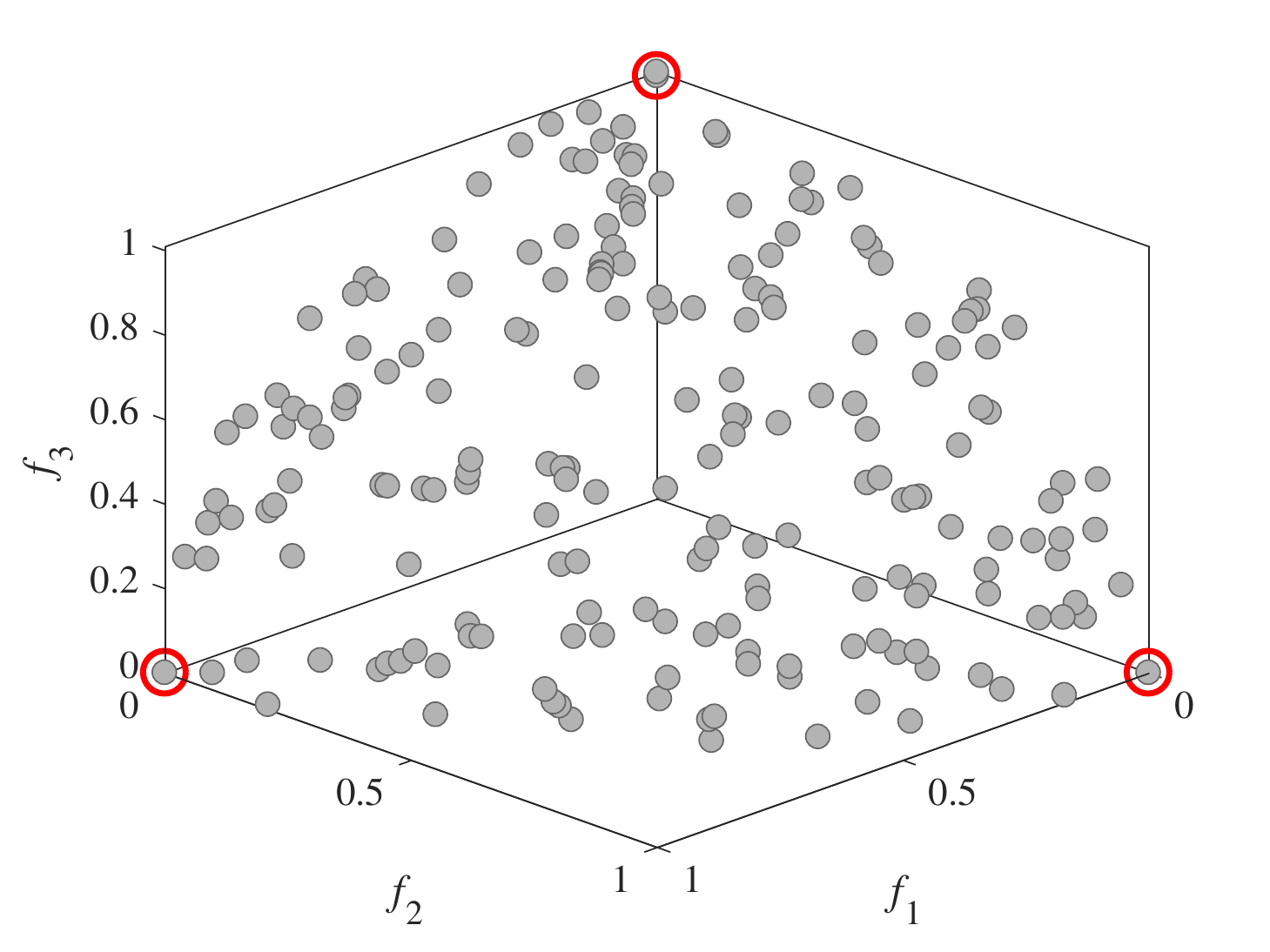}}
  \subfloat[]{
        \label{fig:idtlz2}
      \includegraphics[width=1.8in]{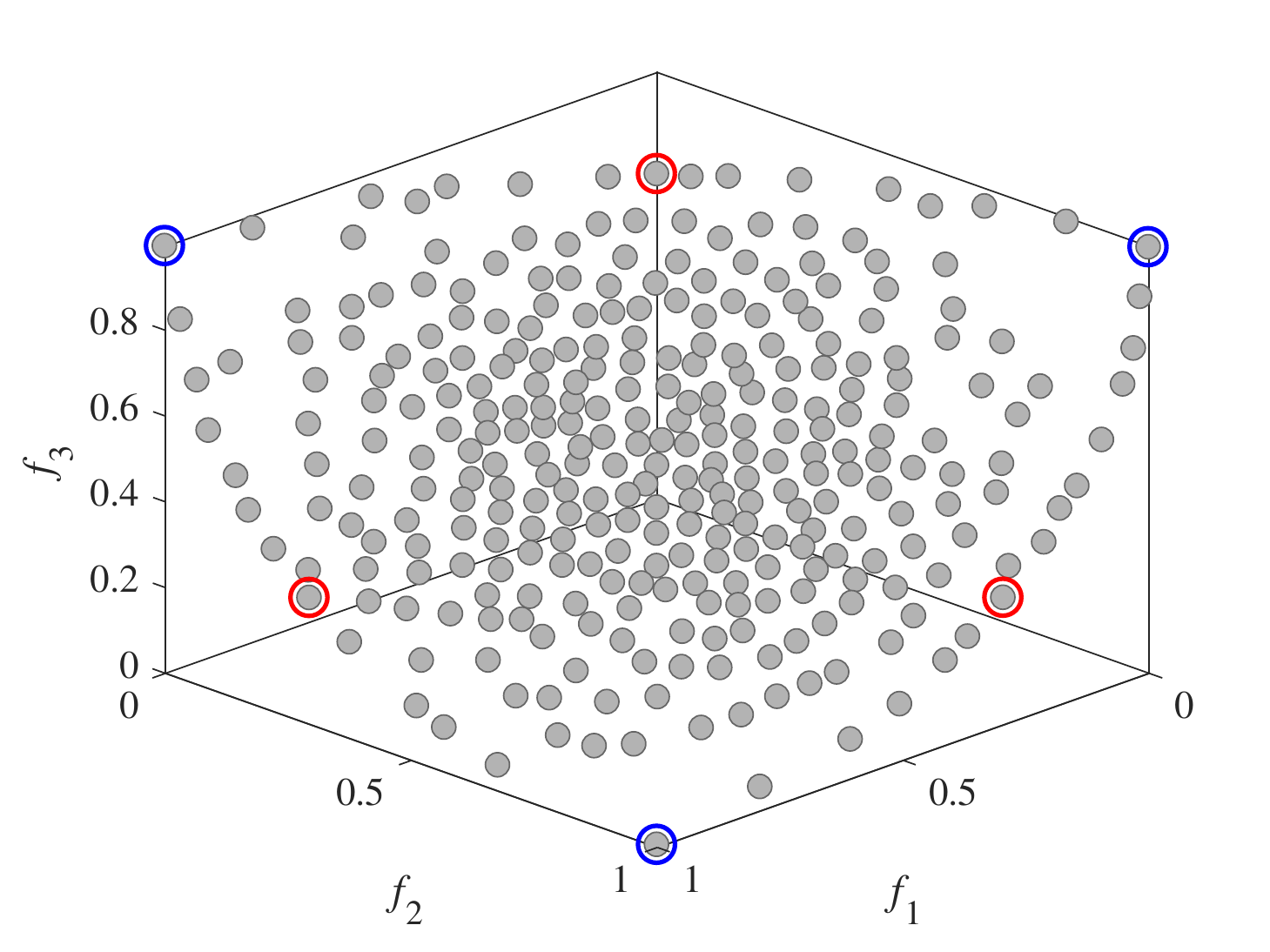}}
  \subfloat[]{
        \label{fig:dtlz5}
      \includegraphics[width=1.8in]{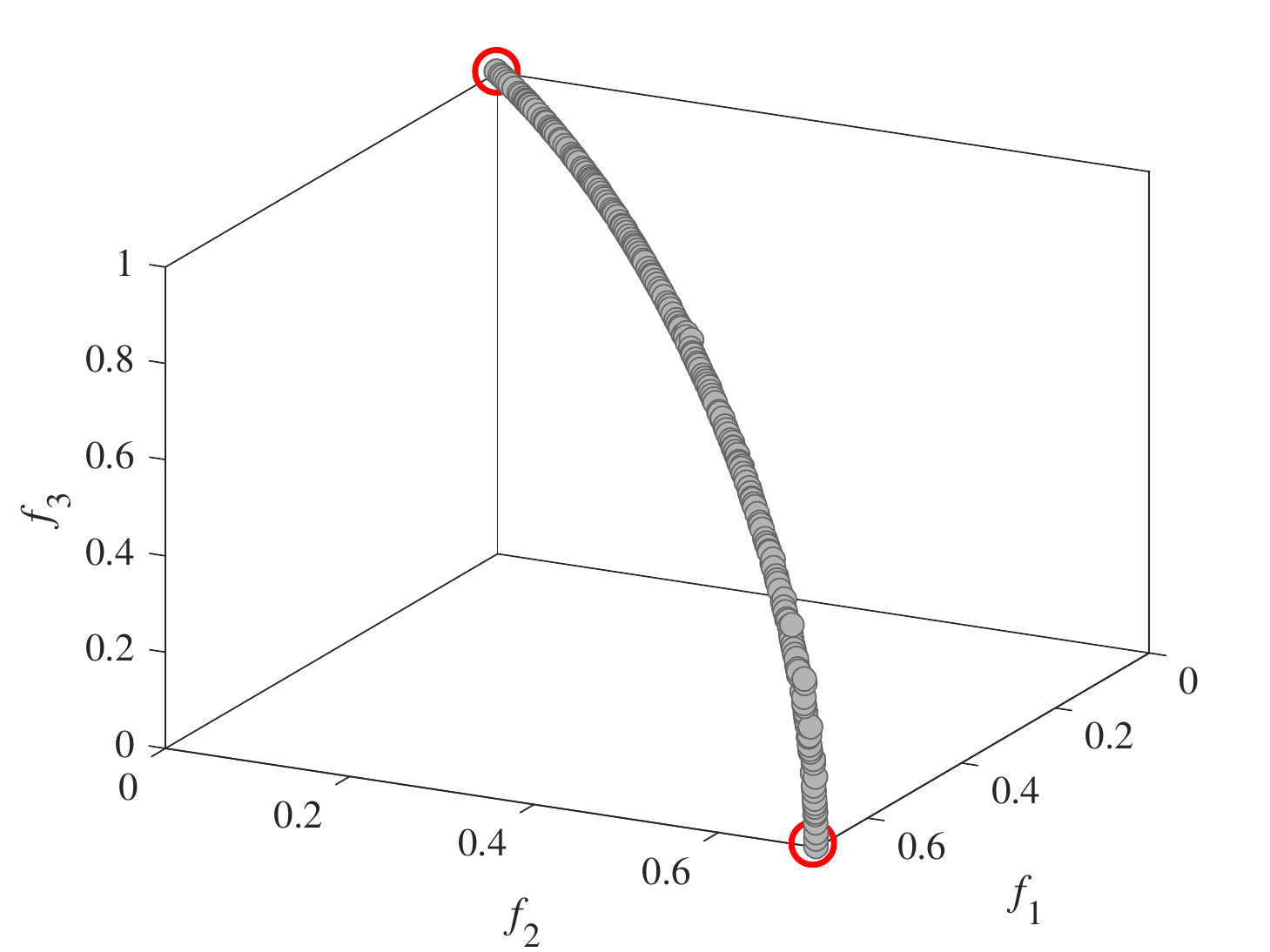}}
  \subfloat[]{
        \label{fig:dtlz7}
      \includegraphics[width=1.8in]{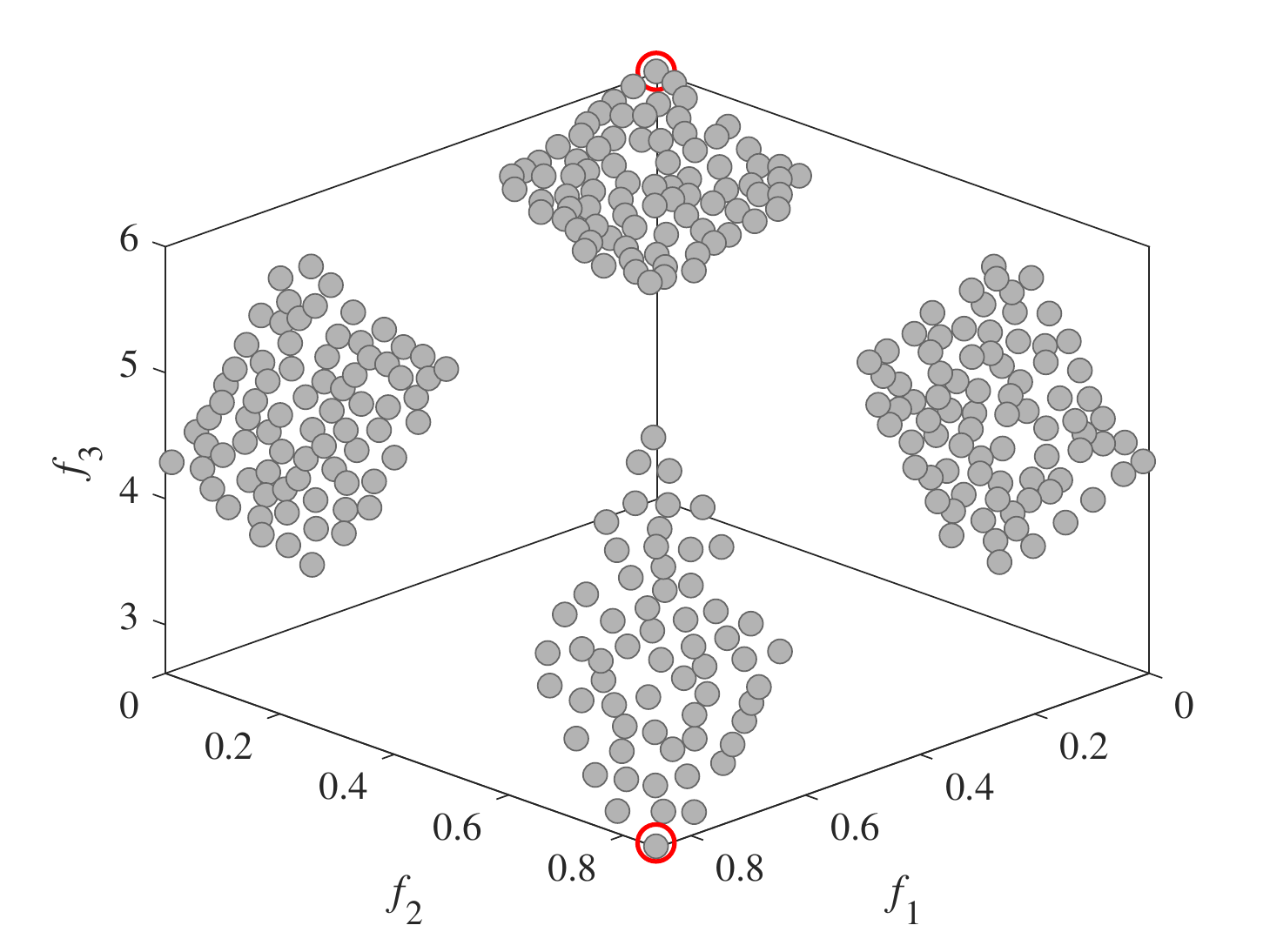}}
  \caption{Four different PF approximations and their corner solutions}\label{fig:corner_solution}
\end{figure*}

\begin{algorithm}
    \caption{Corner Solution Search (\textsc{CS})}\label{alg:cs}
    \Input{
        $P$: a nondominated set;\\
    }
    \Output{
        $P_c$: a set of corner solutions;
    }
    Obtain $P_1$ based on Eq~(\ref{equ:axis});\\
    $P_c = P_1$;\\
    Estimate $z^{nad}$ based on Eq~({\ref{equ:nad}});\\
    Obtain $P_2$ based on Eq~(\ref{equ:plane});\\
    \For{$x \in P_2$}{
        \For {$i = 1$ \KwTo $m$}{
            \If{$f_i(x) > z^{nad}_i$}{
                $P_c = P_c \cup \{x\}$;\\
                \textbf{break};
            }
        }
    }
    \Return $P_c$;
\end{algorithm}

For instance, by using Algorithm \ref{alg:cs}, the corner solutions of four different PFs in Fig. \ref{fig:corner_solution} are approximated (marked in red or blue circles). At first, solutions in $P_1$ are selected first (marked in red circles). Then these solutions are used to approximate $z^{nad}$. However, in Fig. \ref{fig:idtlz2}, as the maximum values of solutions in $P_2$ (marked in blue circles) exceed the values of $z^{nad}$, they are also added to $P_c$. In fact, corner solutions are more important because they contain more information on the shapes of PFs. These solutions are very helpful to approximate the objective limitation of PFs, which can help us locate PFs.

\subsection{The Framework of MaOEA-CS}

The main procedure of MaOEA-CS is presented in Algorithm~\ref{alg:framework}. At first, a population $P$ and the corner solution set $P_c$ are initialized. Then, the reproduction and selection steps are applied to both $P$ and $P_c$ iteratively until the termination criterion is fulfilled. The steps of \textsc{Initialization}, \textsc{Reproduction}, and \textsc{DSA-Selection} in Algorithm \ref{alg:framework} are explained as follows.
\begin{algorithm}
    \caption{Framework of MaOEA-CS (\text{MaOEA-CS})}\label{alg:framework}
    \Input{$N$: Population size;\\
        $\delta$: reproduction probability parameter;\\
        $m$: The number of objectives.
    }
    \Output{The final population.}
    $[P,P_c] = \textsc{Initialization}(N,m)$;\\
    \While{termination criterion is not fulfilled}{
        $Q = \textsc{Reproduction}(P,P_c)$;\\
        $[P,P_c] = \textsc{DSA-Selection}(P\cup Q)$;
    }
    \Return $P$;
\end{algorithm}

\subsection{Initialization}
The initialization steps are given in algorithm \ref{alg:initial}. $m$ direction vectors $E$ along every the coordinate axes are initialized. The population $P$ is randomly generated and then its nondominated solution set is obtained. The corner solution set $P_c$ is also initialized by calling Algorithm \ref{alg:cs}.

\begin{algorithm}
    \caption{Initialization (\textsc{Initialization})}\label{alg:initial}
    \Input{$N$: population size;\\
        $m$: the number of objectives;
    }
    \Output{$P$: The initial population;\\
        $P_c$:cornoer population selected from $P$;
    }
    Initialize $P$ randomly;\\
    \tcc{$E$ stores the direction vector along all the coordinate axes.}
    $E = (e^1,e^2,\dots,e^m)$;\\
    $P = \textsc{nondominated-selection}(P)$;\\
    $P_c = \textsc{CS}(P)$;\\
    \Return $P,P_c$;
\end{algorithm}

\subsection{Reproduction}
Two types of reproduction, called the exploitative search and explorative search, are used in MaOEA-CS. The probability of calling exploitative or explorative search is controlled by a parameter $\delta$. SBX crossover~\cite{Deb1994Simulated} and polynomial mutation~\cite{Deb1999A} operators are adopted as the explorative search. Meanwhile, the mutation operator in~\cite{Liu2014Decomposition,Liu2009The} is adopted as the exploitative search as follows. Given a solution $x$, every component in $x$ is mutated with a probability, $P_m$. If the $i$-th component of $x$ is selected to be mutated, its offspring's $i$-th component $x_i^c$ is computed by:
\begin{equation}\label{equ:mutation}
   \begin{aligned}
    & & &x^i_c = x^i + rnd\times (ub[i]-lb[i]),\\
    & \mbox{where} && rnd = 0.5 \times (rand-0.5) \times (1-rand^{\alpha}),\\
    & & &\alpha = 0.7\times(-(1-\frac{fe}{max\_fe})).
  \end{aligned}
\end{equation}
where $ub[i]$ and $lb[i]$ are the upper and lower bound of $x_i$; $rand$ is a random number in $[0,1]$; $\alpha$ is a simulated annealing variable in which $fe$ is the current number of function evaluations and $max\_fe$ is the maximal allowable number of function evaluations.

The reproduction procedures are presented in Algorithm~\ref{alg:reproduction}. At first, the offspring population $Q$ is initialized to an empty set. When a random number in $[0,1]$ is less than $\delta$, each solution $x$ in $P_c$ undergoes the exploitative search for $\lfloor\frac{|P|}{|P_c|}\rfloor$ times to generate $\lfloor\frac{|P|}{|P_c|}\rfloor$ offspring where $\lfloor\bullet \rfloor$ is the floor function. All the solutions in $P_c$ will generate $N$ offsprings where $N$ is the population size. When the random number in $[0,1]$ is larger than $\delta$, explorative search is conducted on $P$ to generate $N$ offspring.

As the fast convergence towards corner solutions is more important at the early stage and the extension of the corner solutions for diversity becomes more important at the late stage, $\delta$ is switched from a large number for exploitative search in the early stage to a small number ($1-\delta$) for explorative search at the late stage as follows.

\begin{equation}\label{equ:delta-change}
    \Delta_t = \max_{x\in \{1,2,\dots,m\}}\frac{|z^{nad}_i(t) - z^{nad}_i(t-len)|}{|z^{nad}_i(t-len)|}
\end{equation}
where $z^{nad}_i(t)$ indicates the $i$-th objective of $z^{nad}$ approximation at the $t$-the iteration and $len$ is the learning period.

If $\Delta_t$ is less than a preset small number, indicating that exploitative search has already been converged, the value of $\delta$ is switched to ($1-\delta$).

\begin{algorithm}
    \caption{Reproduction (\textsc{Reproduction})}\label{alg:reproduction}
    \Input{$P$: A population;\\
        $P_c$: A corner solution population;
    }
    \Output{$Q$: Offspring population;}
    $Q = \phi$;\\
    \tcc{$\delta$ is a probability parameter to control exploitative and explorative search}
    \eIf{$rand < \delta$}{
        \For{$x \in P_c$}{
            \For{$i=1$ \KwTo $\lfloor\frac{|P|}{|P_c|}\rfloor$}{
                \tcc{Apply exploitative search on $x$ according to Eq~(\ref{equ:mutation})}
                $x^c = \textsc{exploitative-Search}(x)$;\\
                $Q = Q\cup \{x^c\}$;
            }
        }
    }{
        \tcc{Apply SBX crossover and polynomial mutation on all solutions in $P$. }
        $Q = \textsc{explorative-Search}(P)$;
    }
    \Return $Q$;
\end{algorithm}

\subsection{DSA-Selection}
The environmental selection of MaOEA-CS, presented in  Algorithm \ref{alg:dsa}, is called DSA, including \textbf{d}ominance, \textbf{s}pace division and \textbf{a}ngle based selection, detailed as follows.

The nondominated set $R_1$ can be obtained from the population $R$ by calling nondominated selection (line 1). Then, it can be used to approximate ideal point $z^{*}$ as follows.
\begin{equation}\label{equ:ideal}
   \begin{aligned}
    & z^{*} = (z_1^{*},z_2^{*},\dots,z_m^{*})^T,\\
    & \mbox{where~~~} z_i^{*} = \min_{x \in R_1}{f_i(x)}.
  \end{aligned}
\end{equation}

\begin{figure}
  \centering
  \includegraphics[width=3.0in]{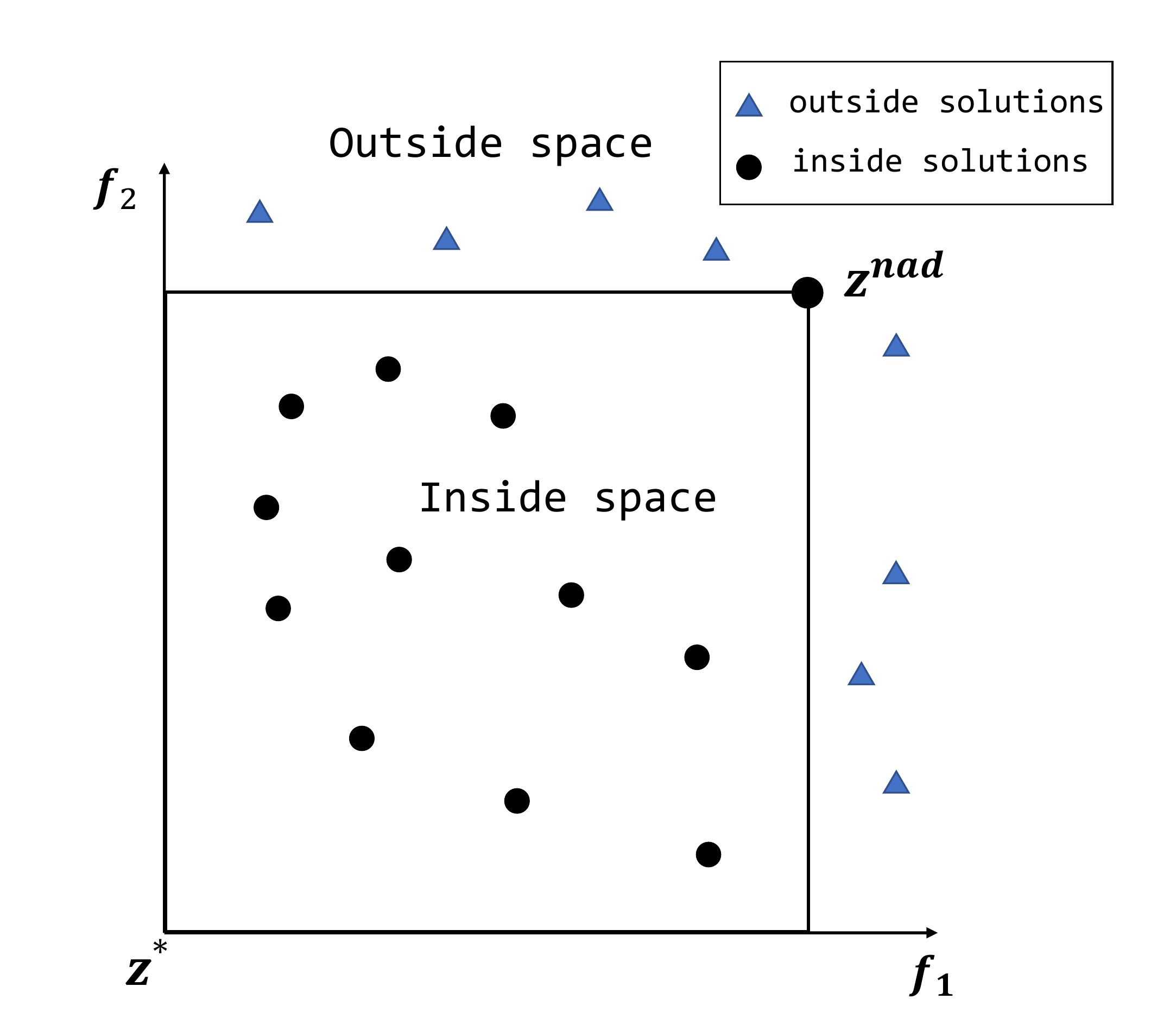}
  \caption{An illustration of division of the inside and outside space by the nadir point approximation.}\label{fig:space}
\end{figure}

The corner solutions are selected from $R_1$ by calling Algorithm \ref{alg:cs} and $z^{nad}$ is computed based on Eq~(\ref{equ:nad}). The objective space can be divided into the inside and outside space by $z^*$ and $z^{nad}$, as shown in Fig. \ref{fig:space}. Given a solution $x$, if there exists $i \in \{1,2,\dots,m\}$, where $f_i(x)$ is larger than $z^{nad}_i$, we say $x$ is located in the outside space. Otherwise, we say $x$ is located in the inside space.

Based on the size of $R_1$, there may three conditions, as follows.
\begin{enumerate}
  \item $|R_1| > N$ (line 5 - 20):  All the solutions in $R_1$ that are located in the inside space are added to $P_{in}$; the rest solutions are added to $P_{out}$. When $|P_{in}| > N$, angle based selection (ABS) (Algorithm \ref{alg:abs}) is called on $P_{in}$ to further select $N$ solutions (line 13 - 14). When $|P_{in}| < N$, $N-|P_{in}|$ solutions closest to $z^{*}$ in $P_{out}$ are selected and added to $P$ with $P_{in}$.
  \item $|R_1| < N$ (line 21 - 23): $N-|R_1|$ solutions nearest to $z^{*}$ are selected from $R\setminus R_1$ and added to $P$ with $R_1$.
  \item $|R_1| = N$ (line 24 - 26): $R_1$ is assigned to $P$ directly.
\end{enumerate}

\begin{algorithm}
    \caption{The selection of MaOEA-CS (\textsc{DSA-Selection})}\label{alg:dsa}
    \Input{$R$: The merged population($|R| > N$);}
    \Output{$P$: The selected solution set ($|P| = N$);\\
        $P_c$: The corner solution set.
    }
    \tcc{Use fast nondominated sorting rank all the solutions.}
    $R_1 = \textsc{Nondominated-Selection}(R)$;\\
    Update $z^*$ based on Eq~(\ref{equ:ideal});\\
    $P_c = \textsc{CS}(R_1)$;\\
    Update $z^{nad}$ based on Eq~(\ref{equ:nad});\\
    \uIf{$|R_1| > N$}{
        $P_{out} = \phi$;\\
        \ForEach{$x \in R_1$}{
            \If{$\exists i \in \{1,2,\dots,m\}, f_i(x) > z^{nad}_i$}{
                $P_{out} = P_{out}\cup\{x\}$;
            }
        }
        $P_{in} = R_1\setminus P_{out}$;\\
        \uIf{$|P_{in}|>N$}{
            $P = \textsc{ABS}(P_{in},P_c,z^*,z^{nad})$;
        }\uElseIf{$|P_{in}|<N$}{
            select $|N-P_{in}|$ solutions closest to $z^*$ from $P_{out}$ and add them to $P_{in}$;\\
            $P = P_{in}$;
        }\Else{
            $P = P_{in}$;\\
        }
    }
    \uElseIf{$|R_1| < N$}{
        Select $|N-R_{1}|$ solutions closet to $z^*$ from $R\setminus R_1$ and add them to $R_{1}$;\\
        $P = R_{1}$;
    }\Else{
        $P = R_1$;
    }
    \Return $P,P_c$;
\end{algorithm}

In the angle-based selection (ABS), the normalized objective vector $F'(x)$ of each solution $x$ can be obtained as follows.
\begin{equation}\label{equ:norm}
   \begin{aligned}
    & & & F'(x) = (f'_1(x),f'_2(x),\dots,f'_m(x))^T,\\
    & \mbox{where} && f'_i(x) = \frac{f_i(x)-z^*_i}{z^{nad}_i-z^*_i}.\\
  \end{aligned}
\end{equation}

The angle between two solutions, $x$ and $y$, can be calculated by:
\begin{equation}\label{equ:label}
  angle(x,y) = \arccos(\frac{F'(x)^T\cdot F'(y)}{\|F'(x)\|\|F'(y)\|})
\end{equation}

The procedures of ABS are presented in Algorithm \ref{alg:abs}. The corner solutions are firstly added to $P$ and deleted from $Q$. For $i$-th solution in $Q$, $\theta_i$ is used to record the minimal angle from it to its nearest solution in $P$. The solution with maximal $\theta$ is added to $P$ and deleted from $Q$ one by one until the population size of $P$ reaches $N$. If the angle between each solution in $Q$ with newly added solution $x$ is larger than its previous $\theta$, $\theta$ is updated by the angle value.

It is worth noting that ABS works similarly to the weight setting method in~\cite{Zhang09b}, except that ABS uses angles to select solutions while distances are used in~\cite{Zhang09b} to select weight vectors (direction vectors).

\begin{algorithm}
    \caption{The angle based selection (\textsc{ABS})}\label{alg:abs}
    \Input{$Q$: A population whose size is larger than $N$;\\
        $P_c$: The corner solution population;\\
        $z^*$: The ideal point;\\
        $z^{nad}$: The nadir point.\\
    }
    \Output{$P$: A population whose size is equal to $N$;}
    $P = \phi$;\\
    $P = P\cup P_c$;\\
    $Q = Q\setminus P_c$;\\
    $\theta = (\theta_1,\theta_2,\dots,\theta_{|Q|})$;\\
    \ForEach{$x^i \in Q$}{
        $\theta_i = \min_{x^j \in P}\{angle(x^i,x^j)\}$;
    }
    \While{$|P|<N$}{
        $k = \argmax_{k\in \{1,2,\dots,|Q|\}}\{\theta_k\}$;\\
        $Q = Q\setminus \{x^k\}$;\\
        $\theta = \theta\setminus \{\theta^k\}$;\\
        $P = P\cup \{x^k\}$;\\
        \ForEach{$\theta_j \in \theta$}{
            $\theta_j = \max\{\theta_j, angle(x^k,x^j)\}$;
        }
    }
    \Return $P$;

\end{algorithm}

\section{Experimental results}
In this section, MaOEA-CS is compared with six state-of-the-art algorithms (BCE-MOEA/D~\cite{Li2016Pareto}, KnEA~\cite{Zhang2015A}, RVEA~\cite{Cheng2016A}, NSGA-III~\cite{Deb2014An}, GSRA~\cite{Chen17G} and RSEA~\cite{He2017A}), in the CEC'2017 Competition on Evolutionary Many-Objective Optimization.

\subsection{Benchmark problems}
Fifteen benchmark functions were proposed in~\cite{cheng2017benchmark} for CEC’2017 competition on evolutionary many-objective optimization. Different from other test suites, the problems in this test suite contain more irregular PFs, aiming to represent various real-world scenarios. The characteristics of all the test instances are summarized in Table \ref{tab:chara}. For each test instance, the number of objectives is set TO $m \in \{5, 10 ,15\}$ respectively and the number of variables is set as suggested in~\cite{cheng2017benchmark}.

\begin{table}
  \centering
  \scriptsize
  \caption{The characteristics of the MaF test suite~\cite{cheng2017benchmark}.}\label{tab:chara}
  \resizebox{3.5in}{!}{
  \begin{tabular}{cc}
    \toprule
    Problem & Characteristics \bigstrut \\
    \midrule
    MaF1 & Linear, No single optimal solution in any subset of objectives \\
    \midrule
    MaF2 & Concave, No single optimal solution in any subset of objectives \\
    \midrule
    MaF3 & Convex, Multimodal \\
    \midrule
    MaF4 & \tabincell{c}{Concave, Multimodal, Badly-scaled, \\ No single optimal solution in any subset of objectives} \\
    \midrule
    MaF5 & Convex, Biased, Badly-scaled \\
    \midrule
    MaF6 & Concave, Degenerate \\
    \midrule
    MaF7 & Mixed, Disconnected, Multimodal \\
    \midrule
    MaF8 & Linear, Degenerate \\
    \midrule
    MaF9 & Linear, Degenerate \\
    \midrule
    MaF10 & Mixed, Biased \\
    \midrule
    MaF11 & Convex, Disconnected, Nonseparable \\
    \midrule
    MaF12 & Concave, Nonseparable, Biased Deceptive \\
    \midrule
    MaF13 & Concave, Unimodal, Nonseparable, Degenerate, Complex Pareto set \\
    \midrule
    MaF14 & Linear, Partially separable, Large scale \\
    \midrule
    MaF15 & Convex, Partially separable, Large scale \\
    \bottomrule
  \end{tabular}}
\end{table}

\subsection{Parameter settings}\label{sec:setting}
The population size, the termination criterion and the running time are set as suggested in~\cite{cheng2017benchmark}. In the reproduction step, $\delta$ is set to $0.9$; the crossover probability $p_c$ is set to $1$; the distribution index $\eta_c$ is set to $20$ for SBX; the mutation probability $p_m$ is set to $1/D$, where $D$ is the number of decision variables; the distribution index $\eta_m$ for polynomial mutation is set to $20$; $\Delta_t$ is set to $0.001\times m$ where $m$ is the number of objectives and the learning period $len$ is set to $50$. The algorithm is programmed in MATLAB and embedded in the open-source MATLAB software platform platEMO~\cite{Tian2017PlatEMO}, suggested by~\cite{cheng2017benchmark}. The parameters in other algorithms are set the same as the original papers.

Each instance was run 31 times. In each run, the maximal allowed number of function evaluations is set to $\max\{100000;10000 \times D\}$ where $D$ is the number of variables. The population sizes of all the compared algorithms are set to $m \times 25$ where $m$ is the number of objectives.

\subsection{Performance Metrics}
IGD~\cite{Sierra2004A, Coello2004A, ZhouZJTO05} and HV~\cite{Zitzler99c} are two widely used performance metrics. Both of them can simultaneously measure the convergence and diversity of the obtained solution set, as follows.
\begin{itemize}
  \item Inverted Generational Distance (IGD): Let $P^*$ be a set of points uniformly sampled over the true PF, and $S$ be the set of solutions obtained by an EMO algorithm. The IGD value of $S$ is computed as:
      \begin{equation}\label{equ:gd}
        IGD(S,P^*) = \frac{\sum_{x\in P^*}dist(x,S)}{|P^*|}
      \end{equation}
      where $dist(x, S)$ is the Euclidean distance between a point $x \in P^*$ and its nearest neighbor in $S$, and $|P^*|$ is the cardinality of $P^*$. The lower is the IGD value, the better is the quality of $S$ for approximating the whole PF.
  \item Hypervolume (HV): Let $r^* = (r_1^*,r_2^*,...,r_m^*)^T$ be a reference point in the objective space that is dominated by all the PF approximation $S$. HV metric measures the size of the objective space dominated by the solutions in $S$ and bounded by $r^*$.
      \begin{equation}\label{equ:hv}
        HV(S)=VOL(\bigcup_{x\in S}[f_1(x),r_1^*] \times...[f_m(x),r_m^*])
      \end{equation}
      where $VOL(\bullet)$ indicates the Lebesgue measure. Hypervolume can measure the approximation in terms of both diversity and convergence. The larger the HV value, the better the quality of PF approximation $S$ is.
\end{itemize}

In this paper, Monte Carlo sampling method~\cite{While2006A} is used to compute HV values. It is worth noting that both IGD and HV values of the nondominated set obtained by all the compared algorithms are computed by platEMO~\cite{Tian2017PlatEMO}.

\subsection{Empirical Results And Discussion}
\begin{table*}
  \centering
  \caption{The performance of seven compared algorithms, in terms of IGD values on MaF1-15.}\label{tab:igd}
  \resizebox{7in}{!}{
\begin{tabular}{ccccccccc}
    \toprule
    Problem & m     & BCE-MOEA/D & GSRA  & KnEA  & RSEA  & RVEA  & NSGA-III & MaOEA-CS \bigstrut\\
    \midrule
    \multirow{3}[2]{*}{MaF1} & 5     & 1.429E-01 (9.3E-04)$^-$ & 1.336E-01 (1.4E-03)$^-$ & 1.246E-01 (2.1E-03)$^-$ & 1.479E-01 (2.9E-03)$^-$ & 3.249E-01 (1.2E-01)$^-$ & 2.073E-01 (1.1E-02)$^-$ & \cellcolor[rgb]{0.749, 0.749, 0.749}1.226E-01 (7.5E-04) \bigstrut[t]\\
          & 10    & 2.555E-01 (1.3E-02)$^-$ & 2.512E-01 (4.7E-03)$^-$ & 2.307E-01 (3.3E-03)$^-$ & 2.425E-01 (4.0E-03)$^-$ & 6.395E-01 (5.9E-02)$^-$ & 2.820E-01 (1.6E-02)$^-$ & \cellcolor[rgb]{0.749, 0.749, 0.749}2.273E-01 (1.3E-03) \\
          & 15    & 3.869E-01 (9.6E-03)$^-$ & 2.925E-01 (6.2E-03)$^-$ & 2.760E-01 (4.7E-03)$^-$ & 2.888E-01 (6.7E-03)$^-$ & 7.000E-01 (7.1E-02)$^-$ & 3.159E-01 (6.9E-03)$^-$ & \cellcolor[rgb]{0.749, 0.749, 0.749}2.536E-01 (1.2E-03) \bigstrut[b]\\
    \midrule
    \multirow{3}[2]{*}{MaF2} & 5     & 1.073E-01 (1.6E-03)$^-$ & 1.145E-01 (2.8E-03)$^-$ & 1.309E-01 (3.6E-03)$^-$ & 1.255E-01 (4.1E-03)$^-$ & 1.279E-01 (1.2E-03)$^-$ & 1.309E-01 (2.7E-03)$^-$ & \cellcolor[rgb]{0.749, 0.749, 0.749}1.006E-01 (1.3E-03) \bigstrut[t]\\
          & 10    & 1.758E-01 (4.8E-03)$^+$ & 1.630E-01 (3.5E-03)$^+$ & \cellcolor[rgb]{0.749, 0.749, 0.749}1.569E-01 (6.6E-03)$^+$ & 3.181E-01 (1.5E-02)$^-$ & 3.670E-01 (2.0E-01)$^-$ & 2.180E-01 (3.3E-02)$^+$ & 2.424E-01 (2.2E-02) \\
          & 15    & 1.976E-01 (1.1E-02)$^+$ & 1.899E-01 (6.0E-03)$^+$ & \cellcolor[rgb]{0.749, 0.749, 0.749}1.812E-01 (4.6E-03)$^+$ & 2.823E-01 (1.4E-02)$^+$ & 6.401E-01 (1.9E-01)$^-$ & 2.100E-01 (8.0E-03)$^+$ & 4.250E-01 (2.0E-02) \bigstrut[b]\\
    \midrule
    \multirow{3}[2]{*}{MaF3} & 5     & 1.347E-01 (1.4E-02)$^-$ & \cellcolor[rgb]{0.749, 0.749, 0.749}7.647E-02 (4.2E-03)$^+$ & 1.597E-01 (7.2E-02)$^-$ & 9.274E-02 (3.4E-02)$^+$ & 9.355E-02 (4.3E-02)$^-$ & 9.743E-02 (1.1E-03)$^+$ & 1.015E-01 (2.4E-03) \bigstrut[t]\\
\bottomrule
    \end{tabular}}
    Wilcoxon's rank sum test at a 0.05 signification level is performed in terms of IGD values. ``$+$",``$-$" or ``$\approx$" indicate that the results obtained by corresponding algorithms is significantly better, worse or similar to that of MaOEA-CS on this test instance, respectively.
\end{table*}

\begin{table*}
  \centering
  \caption{The performance of seven compared algorithms, in terms of HV values, on MaF1-15.}\label{tab:nhv}
   \resizebox{7in}{!}{
    \begin{tabular}{ccccccccc}
    \toprule
    Problem & m     & BCE-MOEA/D & GSRA  & KnEA  & RSEA  & RVEA  & NSGA-III & MaOEA-CS \bigstrut\\
    \midrule
    \multirow{3}[2]{*}{MaF1} & 5     & 9.052E-03 (9.6E-05)$^-$ & 8.700E-03 (1.9E-04)$^-$ & 1.058E-02 (1.8E-04)$^-$ & 9.371E-03 (1.3E-04)$^-$ & 2.269E-03 (8.4E-04)$^-$ & 4.720E-03 (5.4E-04)$^-$ & \cellcolor[rgb]{0.749, 0.749, 0.749}1.104E-02 (1.2E-04) \bigstrut[t]\\
          & 10    & 3.870E-07 (1.4E-07)$^+$ & 2.426E-07 (3.9E-07)$^-$ & 3.793E-07 (1.3E-07)$^+$ & \cellcolor[rgb]{0.749, 0.749, 0.749}5.812E-07 (9.8E-08)$^+$ & 5.963E-09 (2.7E-09)$^-$ & 2.353E-07 (6.2E-08)$^-$ & 3.226E-07 (5.4E-07) \\
          & 15    & 8.063E-13 (9.3E-14)$^+$ & 0.000E+00 (0.0E+00)$^{\approx}$ & 0.000E+00 (0.0E+00)$^{\approx}$ & \cellcolor[rgb]{0.749, 0.749, 0.749}1.229E-11 (1.5E-11)$^+$ & 2.913E-14 (2.4E-14)$^+$ & 3.435E-12 (9.0E-13)$^+$ & 0.000E+00 (0.0E+00) \bigstrut[b]\\
    \midrule
    \multirow{3}[2]{*}{MaF2} & 5     & 1.795E-01 (2.1E-03)$^-$ & \cellcolor[rgb]{0.749, 0.749, 0.749}1.885E-01 (1.7E-03)$^+$ & 1.869E-01 (1.9E-03)$^{\approx}$ & 1.816E-01 (2.2E-03)$^-$ & 1.509E-01 (2.3E-03)$^-$ & 1.516E-01 (4.1E-03)$^-$ & 1.845E-01 (2.4E-03) \bigstrut[t]\\
          & 10    & 1.825E-01 (5.1E-03)$^-$ & 2.197E-01 (3.5E-03)$^+$ & 1.682E-01 (1.2E-02)$^-$ & \cellcolor[rgb]{0.749, 0.749, 0.749}2.382E-01 (2.1E-03)$^+$ & 1.577E-01 (5.1E-02)$^-$ & 2.093E-01 (9.7E-03)$^-$ & 2.127E-01 (5.8E-03) \\
          & 15    & 1.291E-01 (8.5E-03)$^-$ & 2.133E-01 (3.4E-03)$^-$ & 1.063E-01 (1.2E-02)$^-$ & \cellcolor[rgb]{0.749, 0.749, 0.749}2.333E-01 (3.1E-03)$^+$ & 6.223E-02 (2.0E-02)$^-$ & 1.409E-01 (9.6E-03)$^-$ & 2.289E-01 (6.0E-03) \bigstrut[b]\\
    \midrule
    \multirow{3}[2]{*}{MaF3} & 5     & 9.844E-01 (9.6E-03)$^-$ & 9.899E-01 (2.4E-03)$^-$ & 9.534E-01 (6.7E-02)$^-$ & 9.921E-01 (2.7E-02)$^-$ & 9.906E-01 (2.2E-02)$^-$ & 9.987E-01 (9.8E-05)$^{\approx}$ & \cellcolor[rgb]{0.749, 0.749, 0.749}9.988E-01 (8.1E-05) \bigstrut[t]\\
\bottomrule
    \end{tabular}}
    Wilcoxon's rank sum test at a 0.05 signification level is performed in terms of HV values. ``$+$",``$-$" or ``$\approx$" indicate that the results obtained by corresponding algorithms is significantly better, worse or similar to that of MaOEA-CS on this test instance, respectively.
\end{table*}

\begin{table*}
  \centering
  \scriptsize
  \caption{The ranks of all the seven compared algorithms on MaF1-15}\label{tab:rank}
    \resizebox{7in}{!}{
    \begin{tabular}{ccccccccc|ccccccc}
    \toprule
    \multirow{2}[4]{*}{Problem} & \multirow{2}[4]{*}{m} & \multicolumn{7}{c|}{HV}                               & \multicolumn{7}{c}{IGD} \bigstrut\\
    \cmidrule{3-16}      &       & BCE-MOEA/D & GSRA  & KnEA  & RSEA  & RVEA  & NSGA-III & MaOEA-CS & BCE-MOEA/D & GSRA  & KnEA  & RSEA  & RVEA  & NSGA-III & MaOEA-CS \bigstrut\\
    \midrule
    \multirow{3}[1]{*}{MaF1} & 5     & \cellcolor[rgb]{0.988, 0.988, 1.000} 4 & \cellcolor[rgb]{0.792, 0.910, 0.831} 5 & \cellcolor[rgb]{0.976, 0.604, 0.612} 2 & \cellcolor[rgb]{0.980, 0.796, 0.804} 3 & \cellcolor[rgb]{0.388, 0.745, 0.482} 7 & \cellcolor[rgb]{0.592, 0.827, 0.659} 6 & \cellcolor[rgb]{0.973, 0.412, 0.420} 1 & \cellcolor[rgb]{0.988, 0.988, 1.000} 4 & \cellcolor[rgb]{0.980, 0.796, 0.804} 3 & \cellcolor[rgb]{0.976, 0.604, 0.612} 2 & \cellcolor[rgb]{0.792, 0.910, 0.831} 5 & \cellcolor[rgb]{0.388, 0.745, 0.482} 7 & \cellcolor[rgb]{0.592, 0.827, 0.659} 6 & \cellcolor[rgb]{0.973, 0.412, 0.420} 1 \bigstrut[t]\\
          & 10    & \cellcolor[rgb]{0.976, 0.604, 0.612} 2 & \cellcolor[rgb]{0.792, 0.910, 0.831} 5 & \cellcolor[rgb]{0.980, 0.796, 0.804} 3 & \cellcolor[rgb]{0.973, 0.412, 0.420} 1 & \cellcolor[rgb]{0.388, 0.745, 0.482} 7 & \cellcolor[rgb]{0.592, 0.827, 0.659} 6 & \cellcolor[rgb]{0.988, 0.988, 1.000} 4 & \cellcolor[rgb]{0.792, 0.910, 0.831} 5 & \cellcolor[rgb]{0.988, 0.988, 1.000} 4 & \cellcolor[rgb]{0.976, 0.604, 0.612} 2 & \cellcolor[rgb]{0.980, 0.796, 0.804} 3 & \cellcolor[rgb]{0.388, 0.745, 0.482} 7 & \cellcolor[rgb]{0.592, 0.827, 0.659} 6 & \cellcolor[rgb]{0.973, 0.412, 0.420} 1 \\
\bottomrule
    \end{tabular}}%
\end{table*}

Table.~\ref{tab:igd} shows the performance of seven compared algorithms in terms of mean IGD over 31 runs while Table.~\ref{tab:nhv} shows the performance of seven compared algorithms in terms of mean HV over 31 runs. In addition, the overall rankings of seven compared algorithms on all the test problems are presented in Table.~\ref{tab:rank}. It can be observed clearly that the mean rank of all 45 problems for MaOEA-CS in terms of IGD and HV are 2.69 and 2.98, respectively, which won the first ranks among all the seven compared algorithms. This indicates that MaOEA-CS has the best overall performance for MaOPs with different characteristics.

MaF1 has an inverted linear PF. MaOEA-CS performs best on 5-, 10- and 15-objective MaF1, in terms of IGD, among all the compared algorithms.  However, MaOEA-CS performs worse than other compared algorithms, in terms of HV. Especially on 15-objective MaF1, the HV value of the nondominated set obtained by MaOEA-CS is 0. It can be explained as follows. As Monte Carlo sampling method is adopted to compute HV and the true HV for MaF1 (inverted PF) is very small~\cite{Jain2014An, Ishibuchi2016Performance}, the limited sampling in the hypercube can hardly fall into the region determined by the nondominated set and the reference point.

\begin{figure*}
  \centering
  	\subfloat[BCE-MOEA/D]{
		\label{fig:bcemoead}
        \centering
		\includegraphics[width=1.7in]{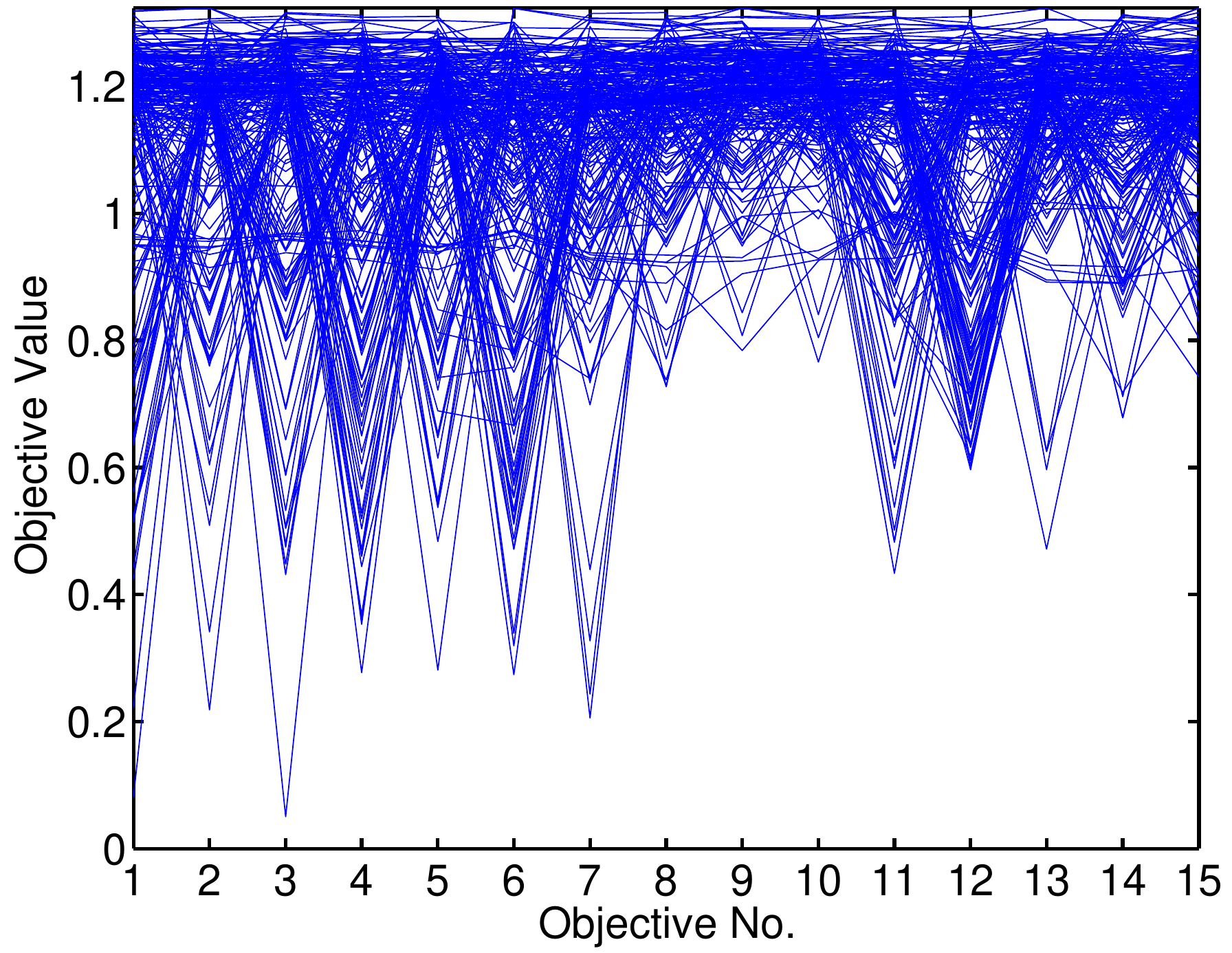}}
	\subfloat[GSRA]{
		\label{fig:gsra}
        \centering
		\includegraphics[width=1.7in]{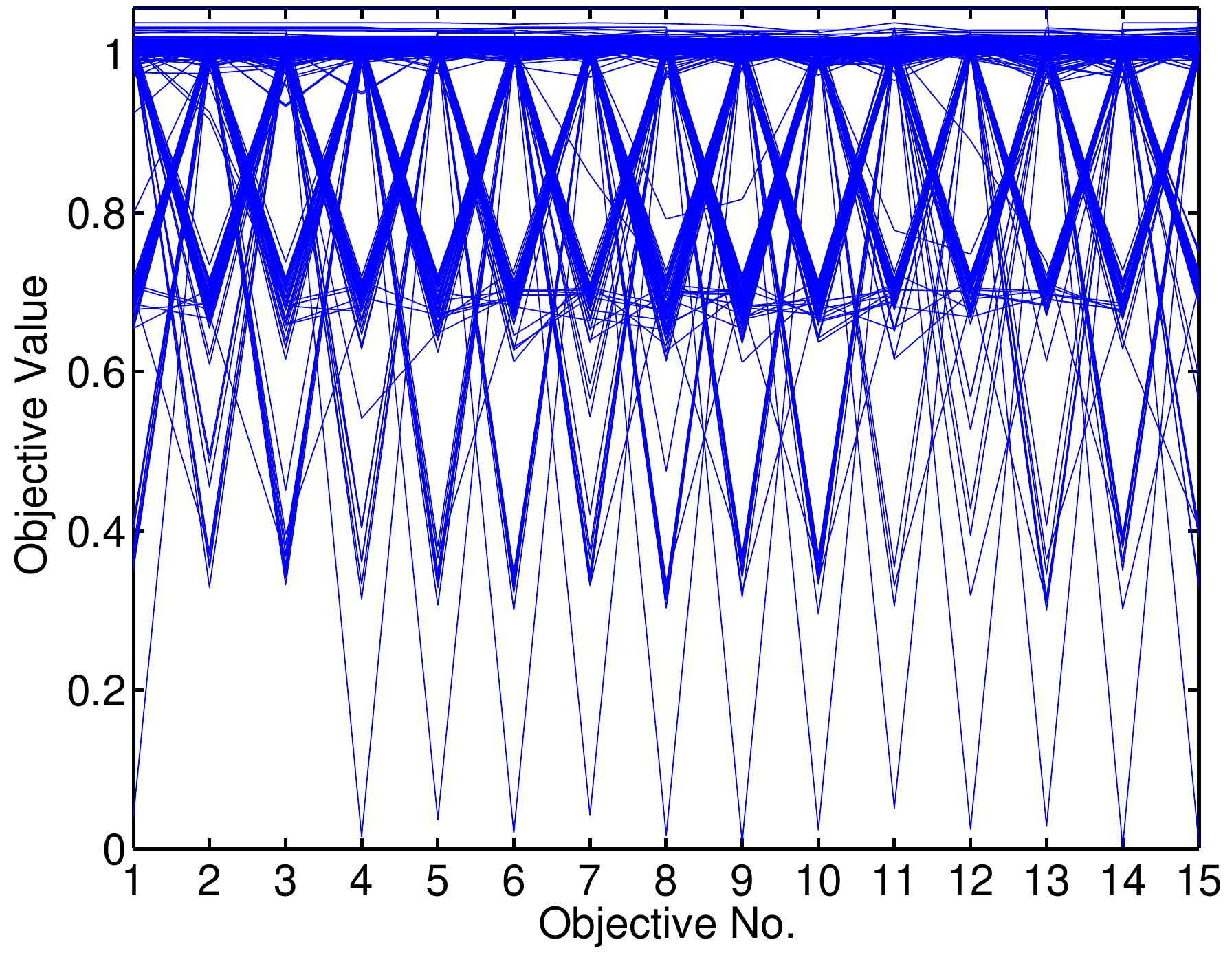}}
    \subfloat[KnEA]{
		\label{fig:knea}
        \centering
		\includegraphics[width=1.7in]{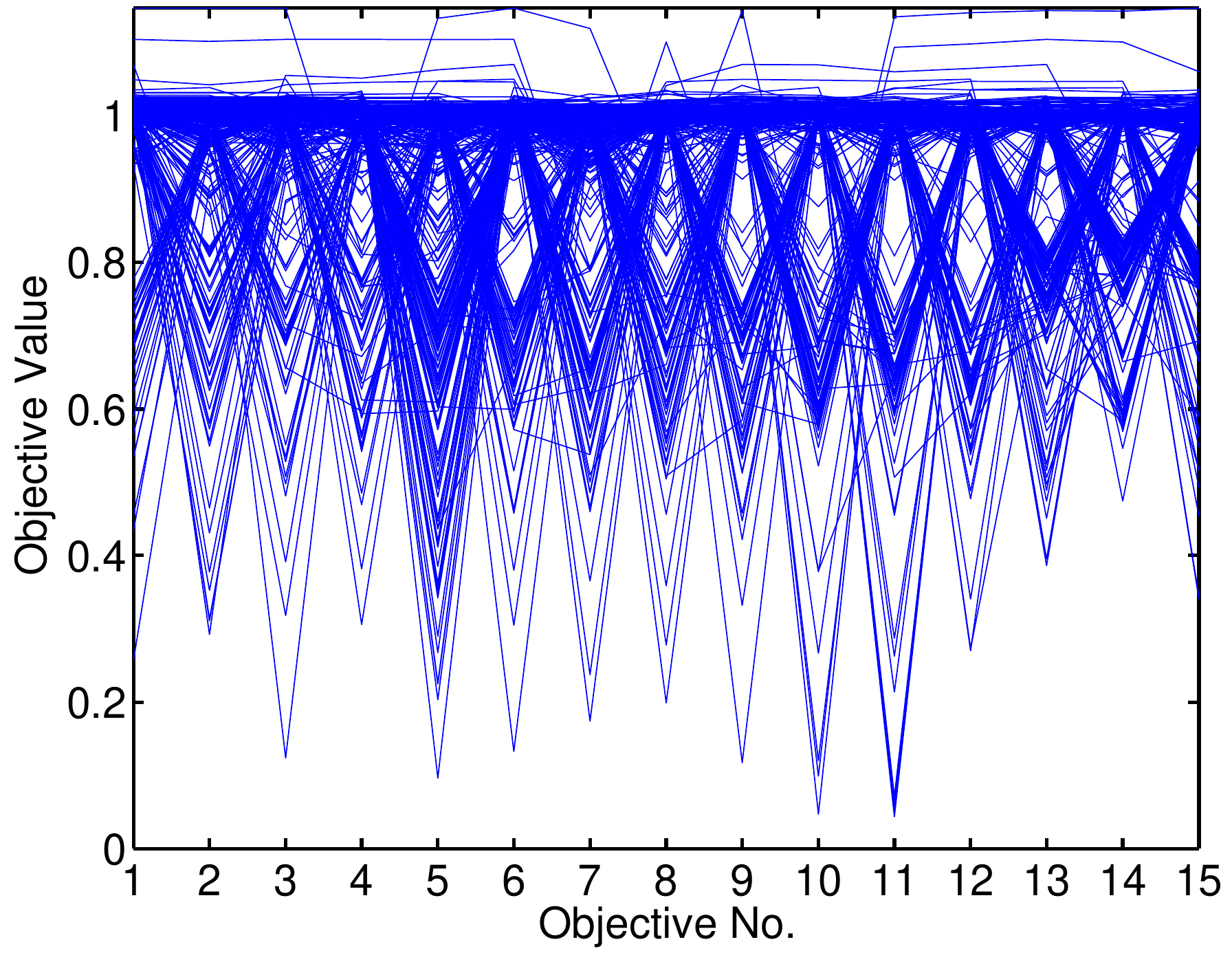}}
    \subfloat[RSEA]{
		\label{fig:rsea}
        \centering
		\includegraphics[width=1.7in]{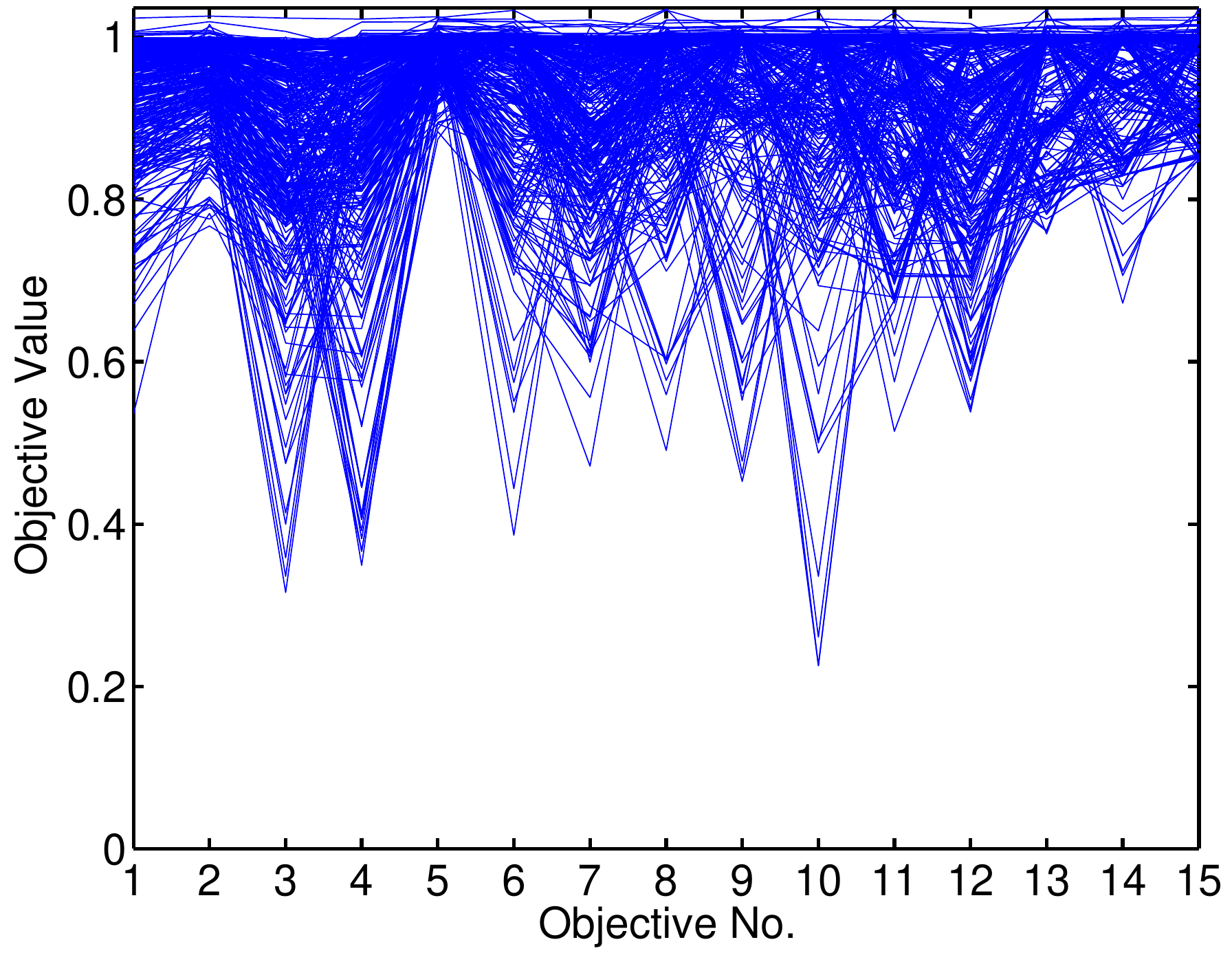}}

    \subfloat[RVEA]{
		\label{fig:rvea}
        \centering
		\includegraphics[width=1.7in]{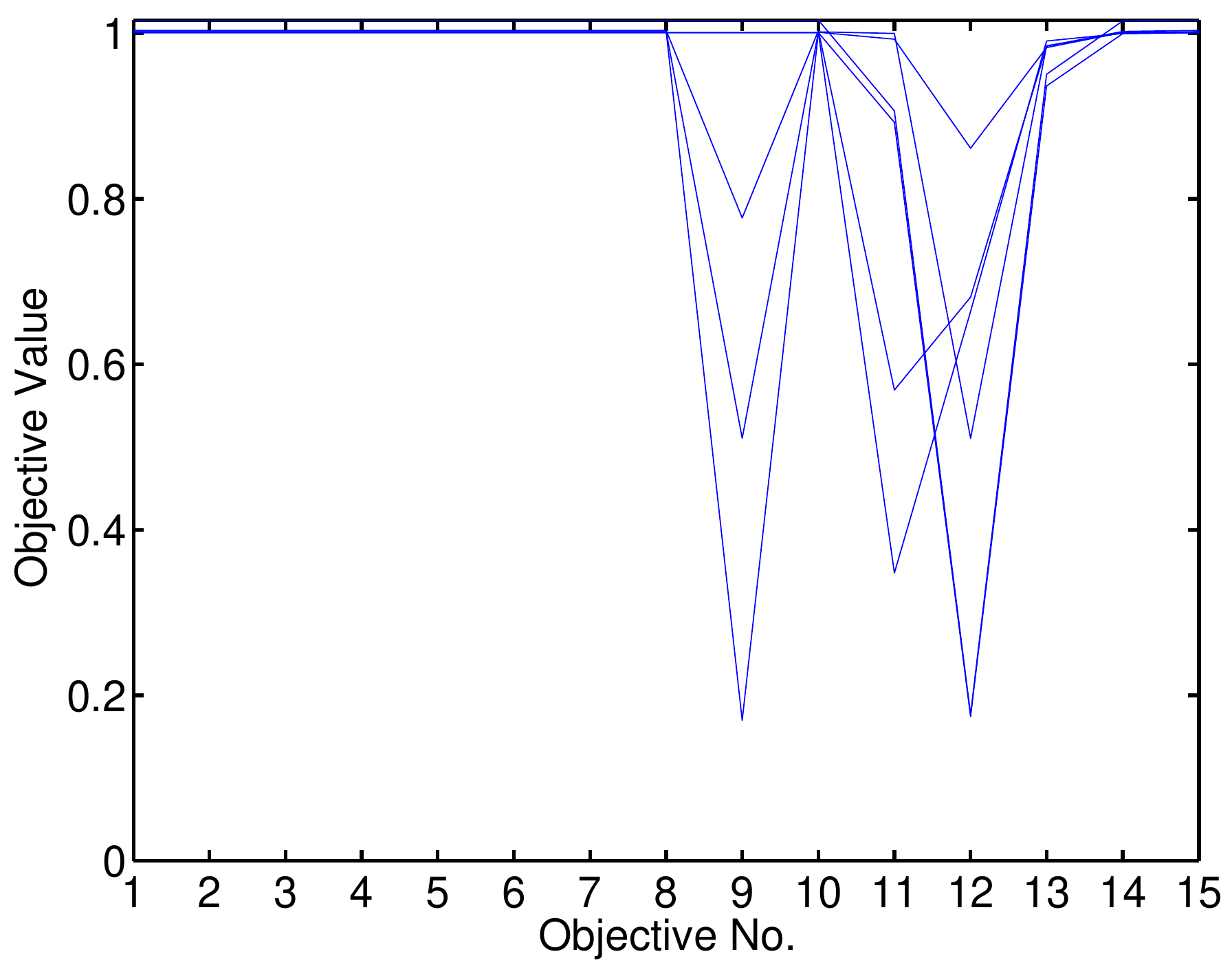}}
    \subfloat[NSGA-III]{
		\label{fig:nsgaiii}
        \centering
		\includegraphics[width=1.7in]{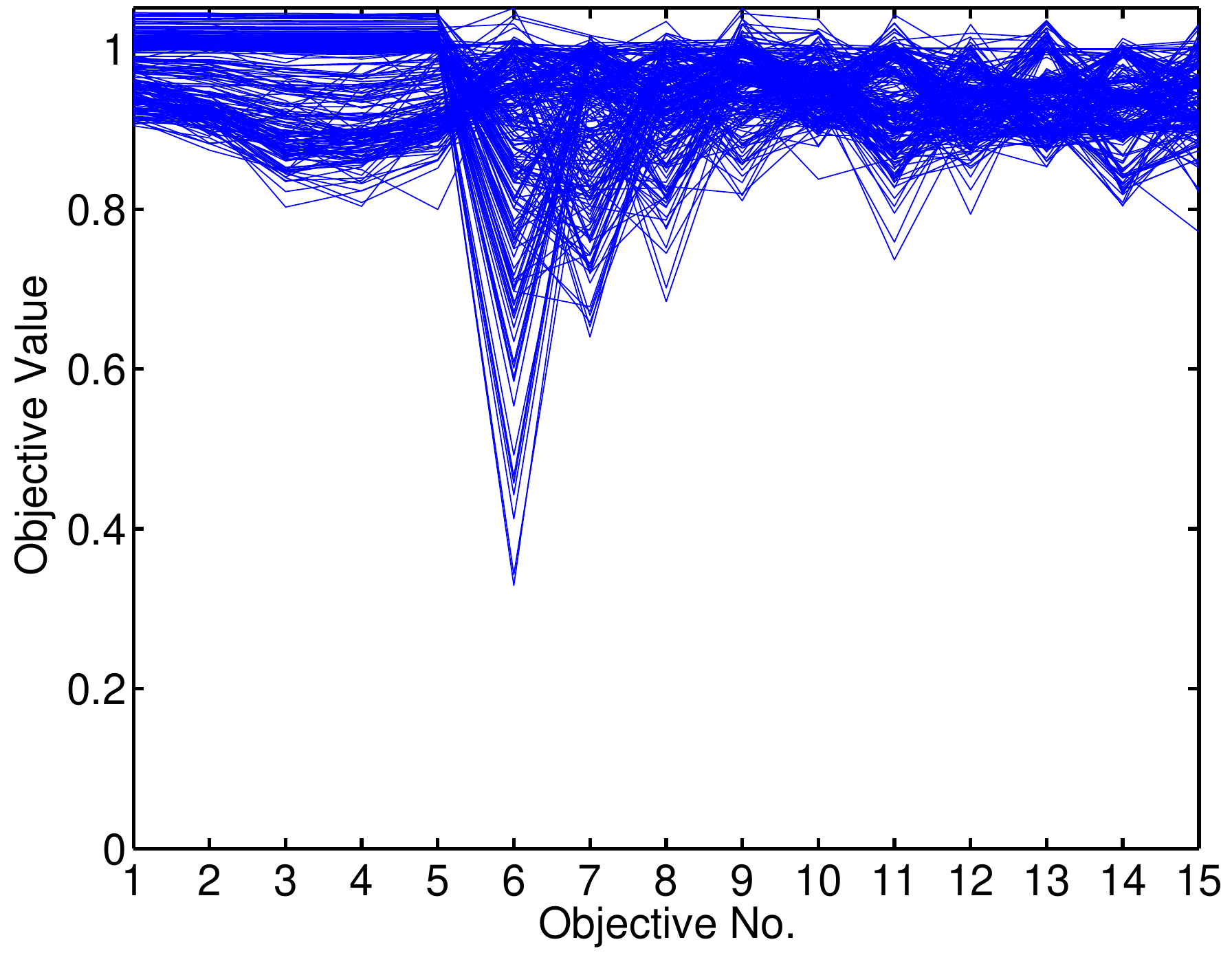}}
    \subfloat[MaOEA-CS]{
		\label{fig:maoeacs}
        \centering
		\includegraphics[width=1.7in]{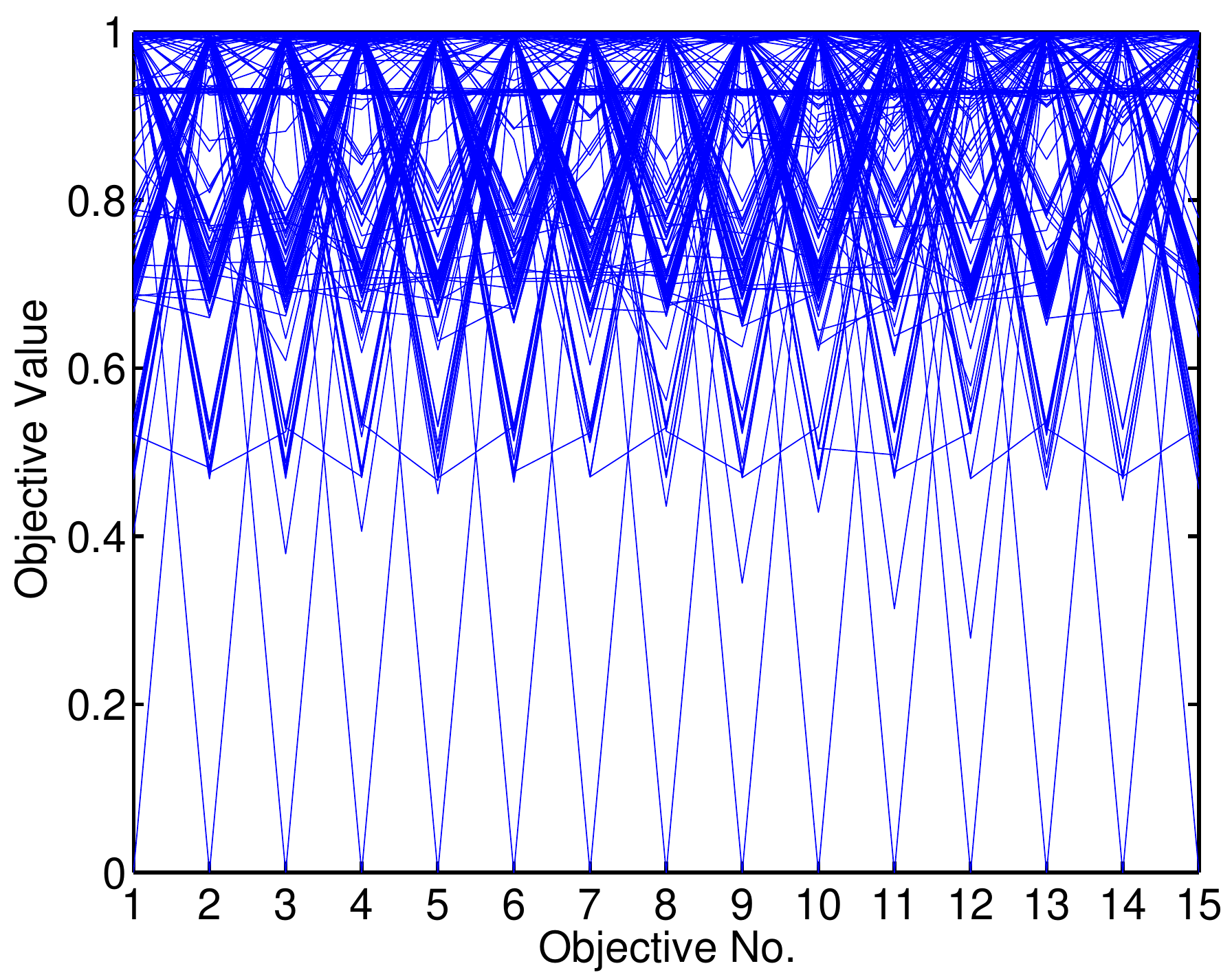}}
  \caption{The parallel coordinates plots of the nondominated sets obtained by seven compared algorithms, in the run with the median IGD values on 15-objective MaF1.}\label{fig:maf1}
\end{figure*}

To further verify the performance of all the compared algorithms in MaF1, the parallel coordinate plots~\cite{Inselberg09} of the nondominated sets obtained by various algorithms on 15-objective MaF1 in the run with the median IGD values are presented in~Fig. \ref{fig:maf1}. It can be observed clearly that MaOEA-CS finds all the boundary solutions and the uniformity of approximation is better than all the other compared algorithms. The convergence of MaOEA-CS is also better than that of other compared algorithms as all the objective values of the solutions obtained by MaOEA-CS are less than 1.0. KnEA, RSEA and GSRA are worse in terms of diversity or convergence. NSGA-III performs worse in terms of diversity and BCE-MOEA/D performs worse in terms of convergence. RVEA is not able to obtain the sufficient number of nondominated solutions.

MaF2 is used to assess whether the MaOEA is able to perform concurrent convergence on different objectives. All the objectives should be optimized simultaneously to well-approximate the true PF. It can be observed from the tables that MaOEA-CS has the best performance in terms of IGD on 5-objective MaF2, while KnEA has the best performance on 10- and 15-objective MaF2. As for HV values, GSRA and RSEA perform better than other compared algorithms.

MaF3 is a multimodel problem with the convex PF. This problem is mainly to test whether the MaOEA can deal with convex PFs. It can be observed from the tables that MaOEA-CS has the best performance on MaF3 (5, 10, 15-objectives), in terms of HV, while GSRA has the best performance in terms of IGD.

MaF4 is also a multimodel problem with THE convex PF, but it is more difficult than MaF3 as it is not well-scaled. It can be observed from the tables that MaOEA-CS performs best on all MaF4 problems, in terms of IGD while it only has the best performance on 5-objective MaF4 in terms of HV. NSGA-III and RSEA have the best performance, in terms of HV, on 10- and 15-objective MaF3, respectively.

The PS of MaF5 has a highly biased distribution, where the majority of Pareto optimal solutions are crowded in a small subregion. In addition, MaF5 is also a badly-scaled problem with objective values ranging from 1 to 1024. It can be observed from the tables that MaOEA-CS has the best performance in terms of IGD, while NSGA-III have the best performance, in terms of HV, on 10-objective MaF5 and KnEA has the best performance in terms of HV, on 15-objective MaF5.

MaF6 is a degenerated problem whose PF is a 2-dimensional manifold regardless of the number of objectives. It can be observed from the tables that MaOEA-CS has the best performance in terms of IGD and HV on all the MaF6 with different number of objectives, except for 10-objective MaF6, where BCE-MOEA/D has the best performance in terms of IGD.

MaF7 has a disconnected PF where the number of disconnected segments is $2^m-1$ ($m$ is the number of objectives). MaF7 can be used to test whether an MaOEA can handle MaOPs with the disconnected PFs. It can be observed from the tables that GSEA and KnEA performs better in terms of IGD and RSEA performs better in terms of HV. MaOEA-CS performs well in terms of both IGD and HV, although it does not have the best performance.

Both MaF8 and MaF9 have two-dimensional decision space. MaF8 calculates the Euclidean distance from a given point $x$ to a set of $M$ target points of a given polygon while MaF9 calculates the Euclidean distance from $x$ to a set of $M$ target straight lines, each of which passes through an edge of the given regular polygon with $M$ vertexes. It can be observed from Table.~\ref{tab:rank} that, although MaOEA-CS does not obtain the best performance, it always ranks the first or second among all the seven compared algorithms. This indicates that MaOEA-CS has more stable performance on MaF8-9.

MaF10 has a scaled PF containing both convex and concave segments, which is used to test whether the algorithm can handle PFs of complicated and mixed geometries. It can be observed from the tables that NSGA-III and RVEA perform better in terms of IGD; while NSGA-III performs better in terms of HV.

MaF11 is used to assess whether an MaOEA is capable of handle an MaOP with the scaled and disconnected PFs. It can be observed from the tables that BCE-MOEA/D and MaOEA-CS perform better in terms of IGD; while NSGA-III and RSEA performs better in terms of HV.

As for MaF12, its decision variables are nonseparably reduced, and its fitness landscape is highly multimodal. It can be observed from the tables that MaOEA-CS has the best performance in terms of IGD; while KnEA has the best performance in terms of HV.

MaF13 has a simple concave PF that is always a unit sphere regardless of the number of objectives. However, its decision variables are nonlinearly linked with the first and second decision variables, thus leading to difficulty in convergence. This problem is used to test whether the algorithm can handle an MaOP with the complicated PS. GSRA has the best performance in terms of both IGD and HV.

MaF14 and MaF15 are mainly used to assess whether an MaOEA can handle complicated fitness landscape with mixed variable separability, especially in large-scale cases. MaOEA-CS has very good overall performance as it can be observed from Table.~\ref{tab:rank} that MaOEA-CS ranks either first or second among all the seven compared algorithms in terms of HV or IGD.

\subsection{Parameter Sensitivity Studies}
Two parameters, i.e., the switching threshold $\Delta_t$ and the learning period $len$, exist in MaOEA-CS. In this section, the sensitivity of them with regard to MaOEA-CS is investigated. In the experiments, $\Delta_t$ is set to 1.0, 0.1, 0.01, 0.001, 0.0001 or 0.00001 and $len$ is set from 10 to 90 with the step size 20. The experiments are conducted on 10-objective MaF4 and MaF9 with a total number of $6 \times 5$ different parameter configurations. 30 independent runs have been conducted for each configuration on each test problem.

\begin{figure}
  \centering
  \includegraphics[width=3.0in]{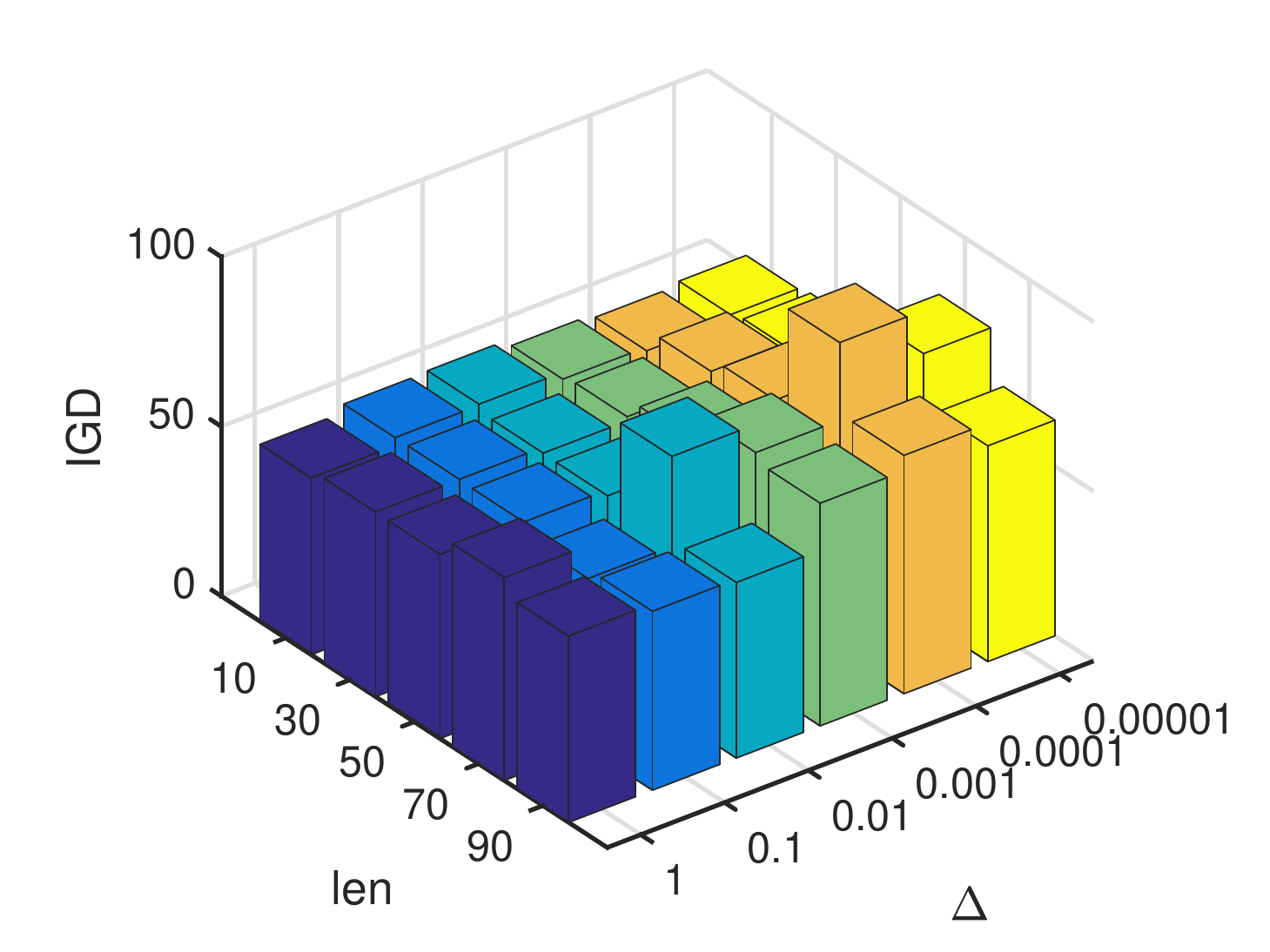}
  \caption{Mean IGD values obtained by MaOEA-CS with 30 different combinations of $\Delta_t$ and $len$ on 10-objective MaF4}\label{fig:maf4-para}
\end{figure}

Fig. \ref{fig:maf4-para} shows the IGD values obtained by MaOEA-CS with 30 different combinations of $\Delta_t$ and $len$ on 10-objective MaF4. It can be observed that, MaOEA-CS achieves the good performance in terms of IGD when the setting of learning period $len$ positively correlates to the setting of the switching threshold $\Delta_t$. This can be explained as follows.

In MaOEA-CS, the first search process (emphasizing exploitative search) should be conducted until the corner solutions have been approximated (i.e., nadir point do not change much). After that, the second search process (emphasizing explorative search) is conducted for improving the diversity of the solution set. If the learning period $len$ is set to a small value, the change of the nadir point also tends to be a small value. In other words, a small value of the learning period $len$ also requires a small value of switching threshold $\Delta_t$, and vice versa. In addition, it also can be observed from Fig. \ref{fig:maf4-para} that the best performance can be achieved in terms of IGD on MaF4, when $\Delta_t = 0.1$ and $len = 70$.

\begin{figure}
  \centering
  \includegraphics[width=3.0in]{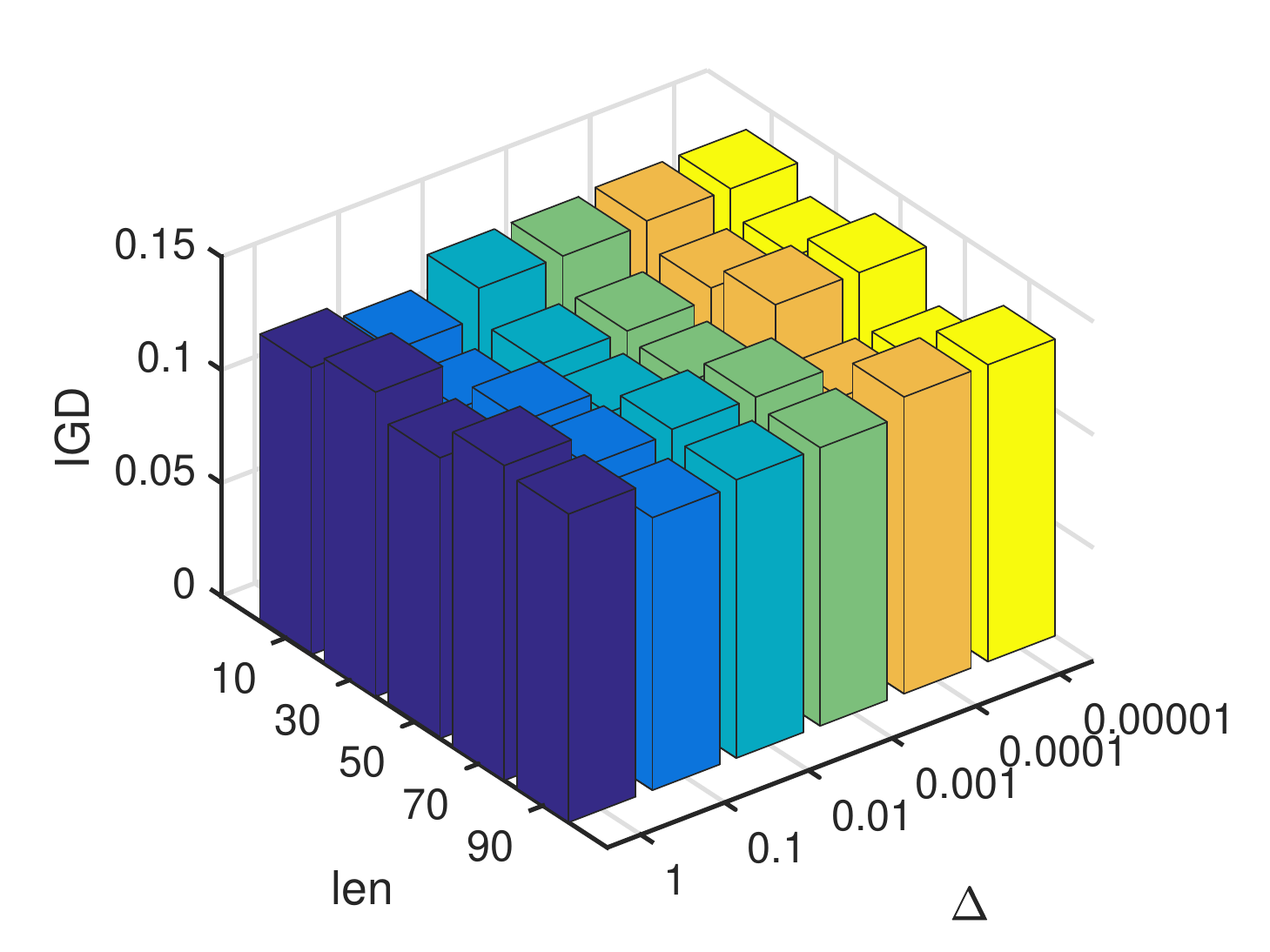}
  \caption{Mean IGD values obtained by MaOEA-CS with 30 different combinations of $\Delta_t$ and $len$ on 10-objective MaF9}\label{fig:maf9-para}
\end{figure}

Fig. \ref{fig:maf9-para} shows the IGD values with 30 different parameters on 10-objective MaF9. Similar phenomenon can be observed that the setting of $len$ and $\Delta_t$ should be positively correlated to achieve good performance. It also can be observed from Fig. \ref{fig:maf9-para} that the best performance can be achieved in terms of IGD on MaF9, when $\Delta_t = 0.1$ and $len = 30$.

%
%
%

\section{Applications on the real-world optimization problems}
In this section, MaOEA-CS is applied and compared with other six algorithms on two real-world engineering optimization problems.
\begin{enumerate}
  \item Crash-worthiness design of vehicles (CWDV) can be formulated as the structural optimization on the frontal structure of vehicle for crash-worthiness~\cite{Liao2007MULTI}. Thickness of five reinforced members around the frontal structure are chosen as the design variables, while mass of vehicle, deceleration during the full frontal crash and toe board intrusion in the offset-frontal crash are considered as three objectives. More detailed mathematical formulation can be found in~\cite{Liao2007MULTI}.
  \item Car side-impact problem (CSIP) aims at finding a design that balances between the weight and the safety performance. It is firstly formulated for the minimization of the weight of the car subject to some safety restrictions on safety performance~\cite{Gu2001Optimisation,Deb2009Reliability}. In~\cite{Deb2009Reliability}, it is reformulated as a 9-objective optimization problem by treating some constraints as objectives. More details of the mathematical formulation can be found in~\cite{Deb2009Reliability}.
\end{enumerate}

\subsection{Experimental setups}
For CWDV, the population size is set to 120; and the maximum number of allowed iterations is set to 200. For CSIP, The population size are set to 210; and the maximum number of iterations is set to 2000. All the seven compared algorithms are run for 30 times on each problem. Other parameters in all the compared algorithm are set the same as that in Section. \ref{sec:setting}.

As the real PFs of both CWDV and CSIP are unknown, to compute IGD, a reference PF (denoted as $P^*$) is constructed by obtaining all the nondominated solutions of all 30 runs obtained by all the compared algorithms for each problem. To compute HV, each objective value of all the nondominated solutions are firstly normalized to [0, 1] by the maximal and minimal values of $P^*$; and the reference point is set to $(1.1, \ldots, 1.1)^T$.

\subsection{Experimental results}
\begin{figure*}
  \centering
  	\subfloat[BCE-MOEA/D]{
		\label{fig:cwdp.bcemoead}
        \centering
		\includegraphics[width=1.7in]{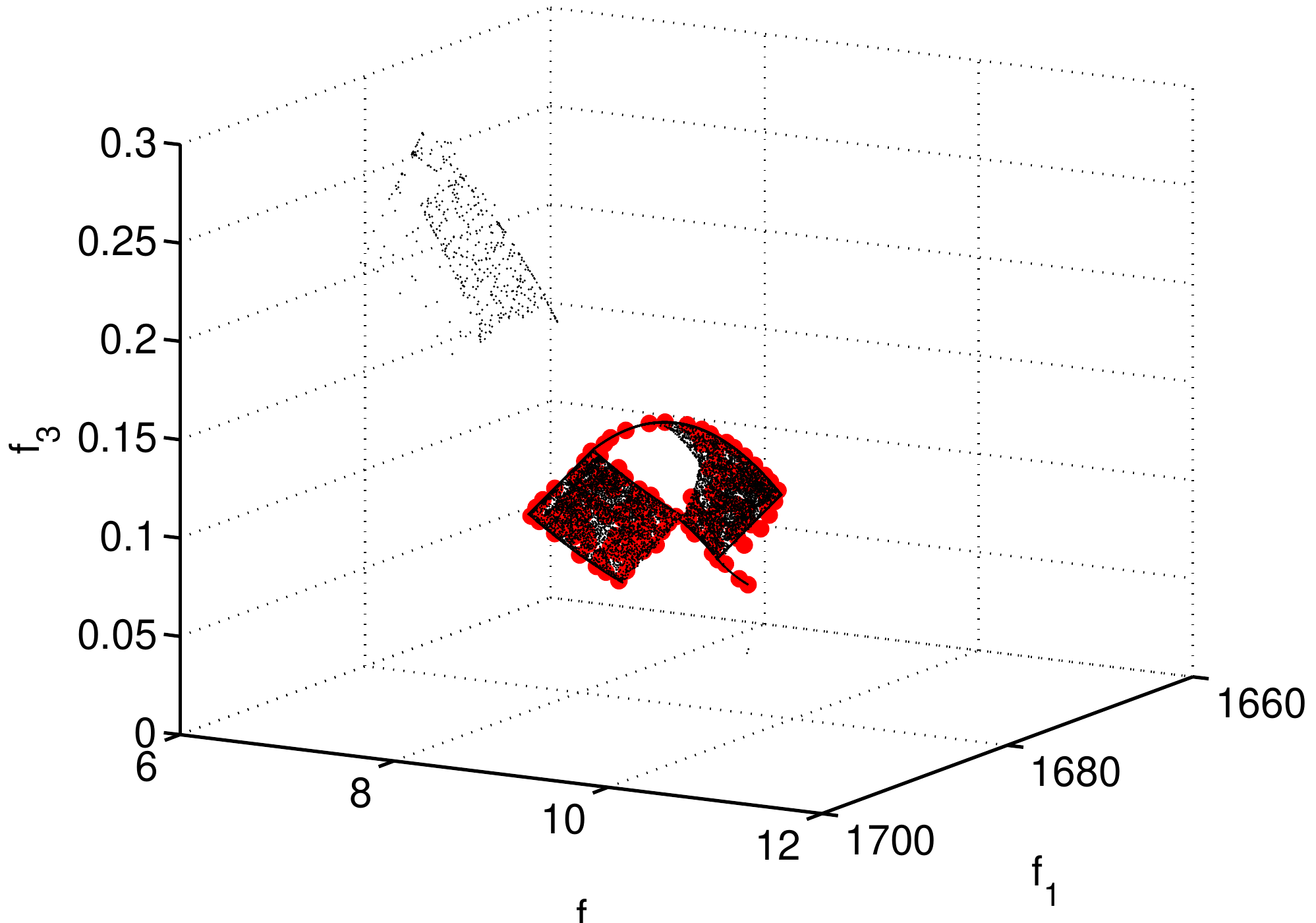}}
	\subfloat[GSRA]{
		\label{fig:cwdp.gsra}
        \centering
		\includegraphics[width=1.7in]{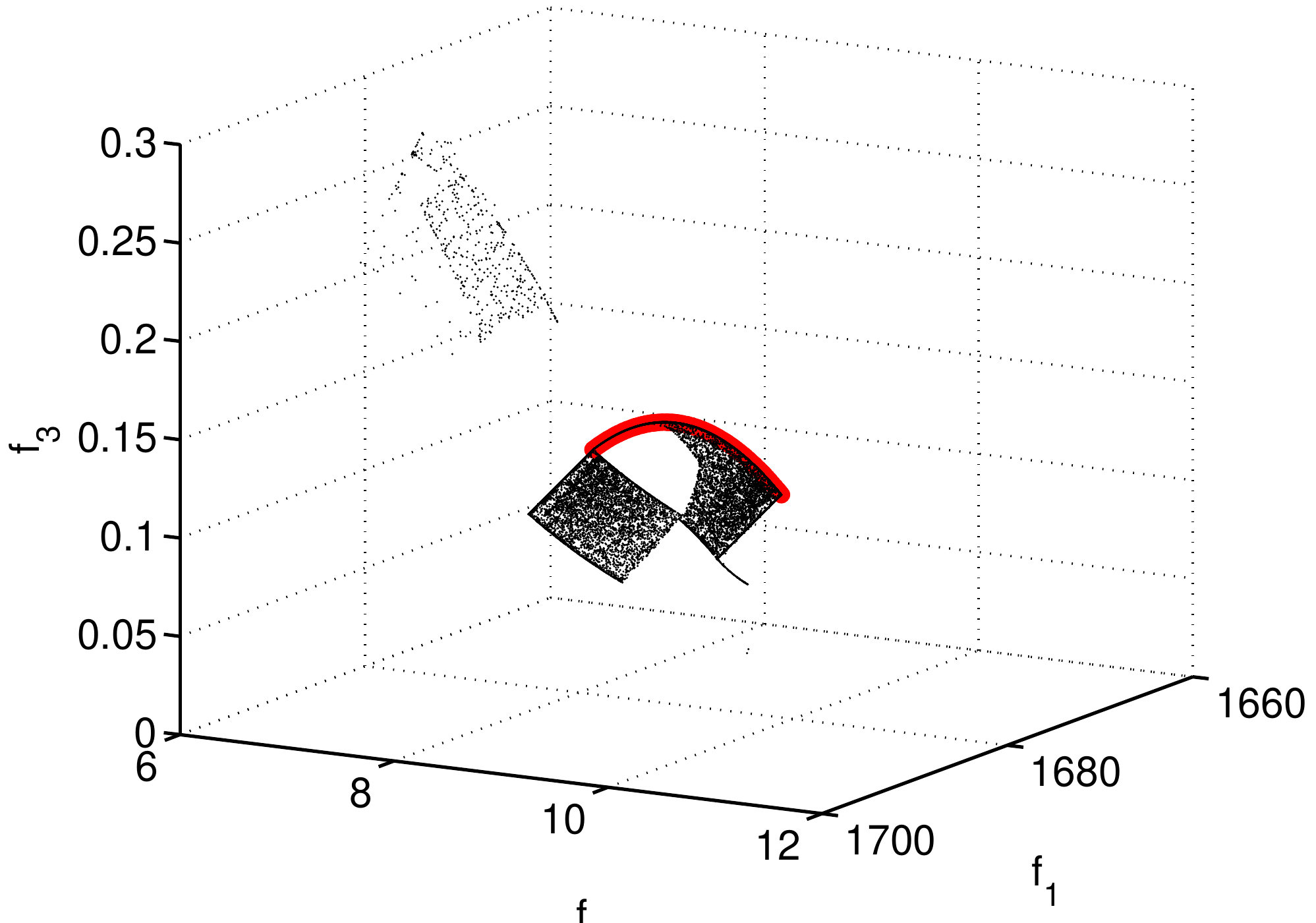}}
    \subfloat[KnEA]{
		\label{fig:cwdp.knea}
        \centering
		\includegraphics[width=1.7in]{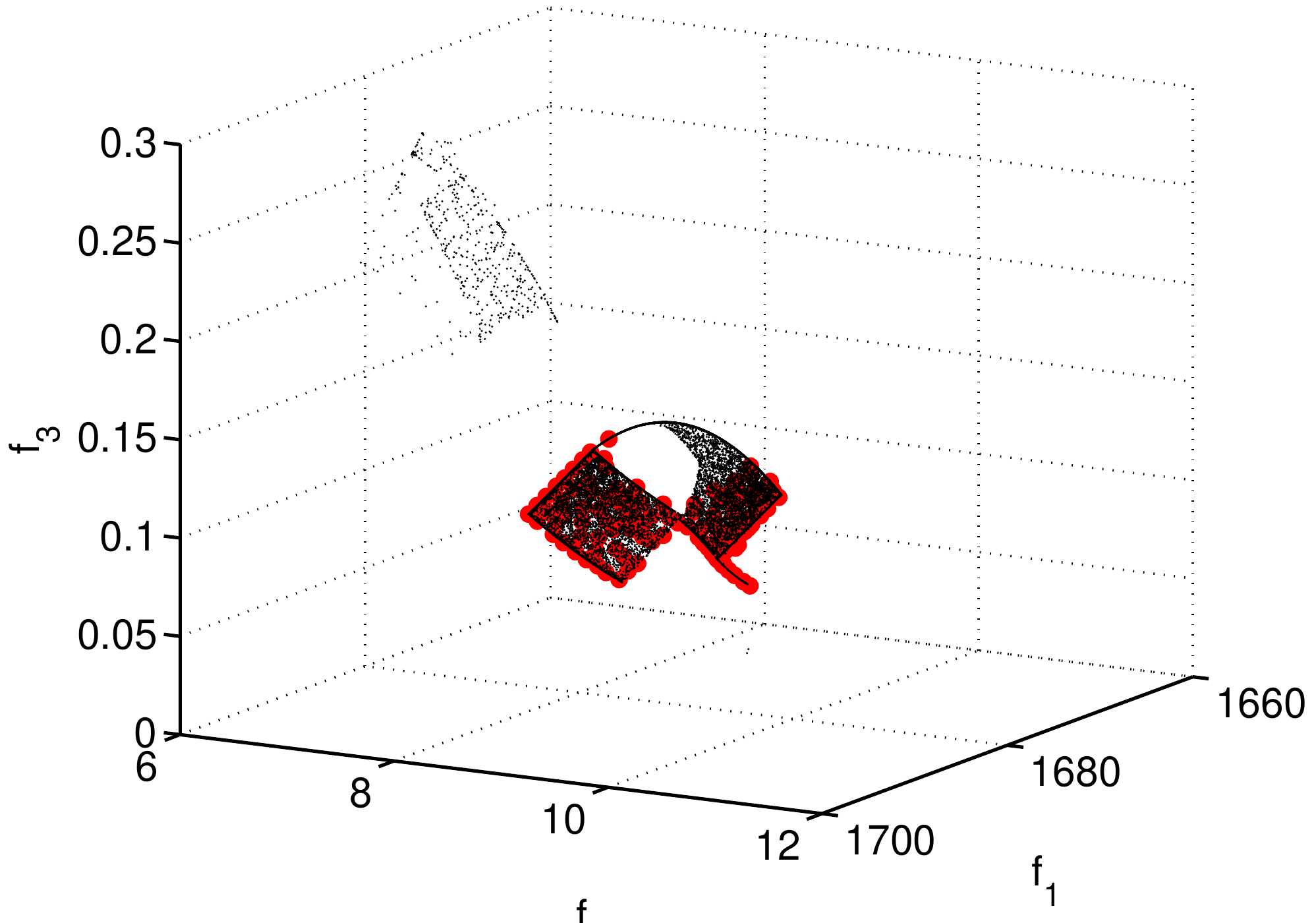}}
    \subfloat[RSEA]{
		\label{fig:cwdp.rsea}
        \centering
		\includegraphics[width=1.7in]{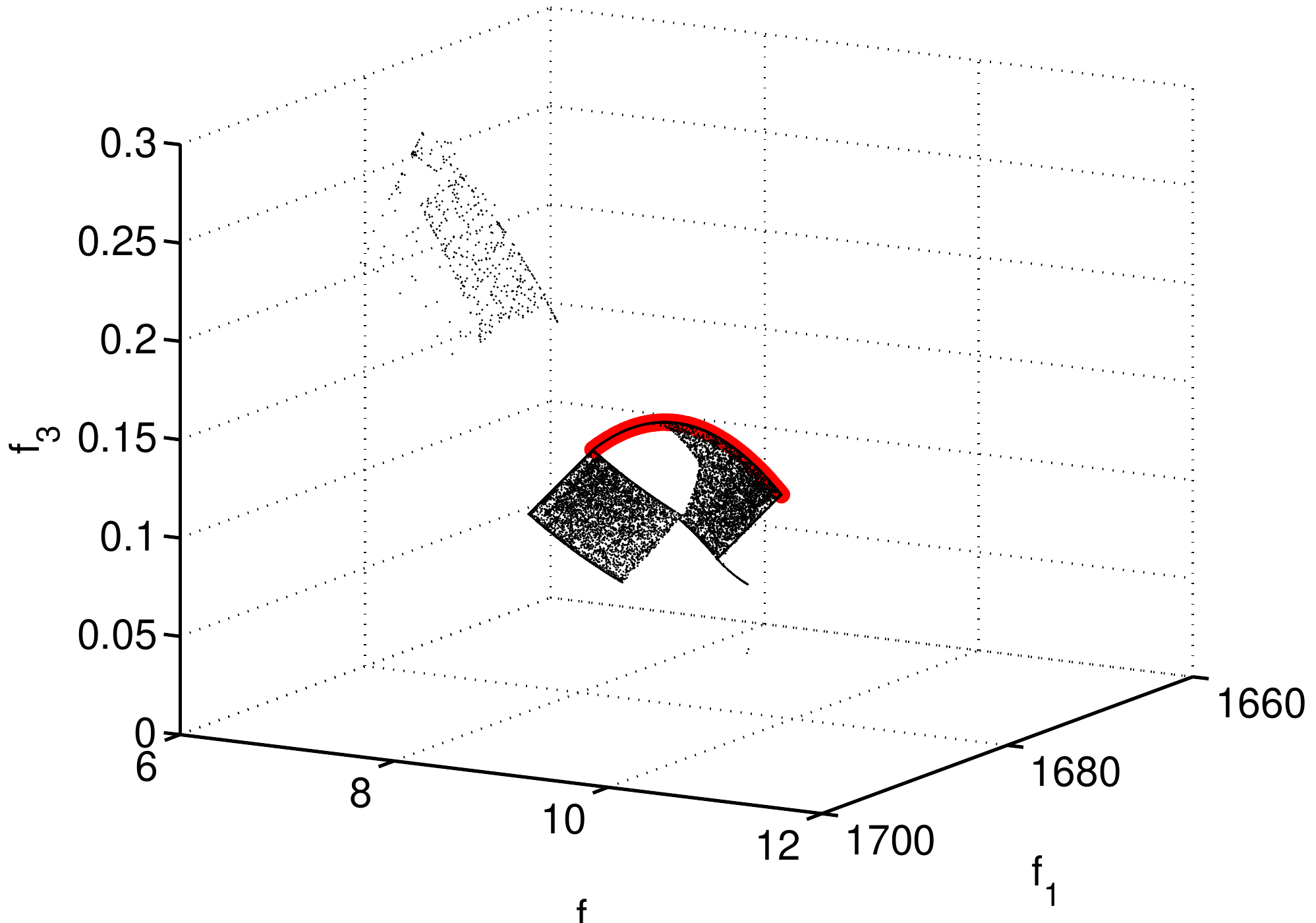}}

    \subfloat[RVEA]{
		\label{fig:cwdp.rvea}
        \centering
		\includegraphics[width=1.7in]{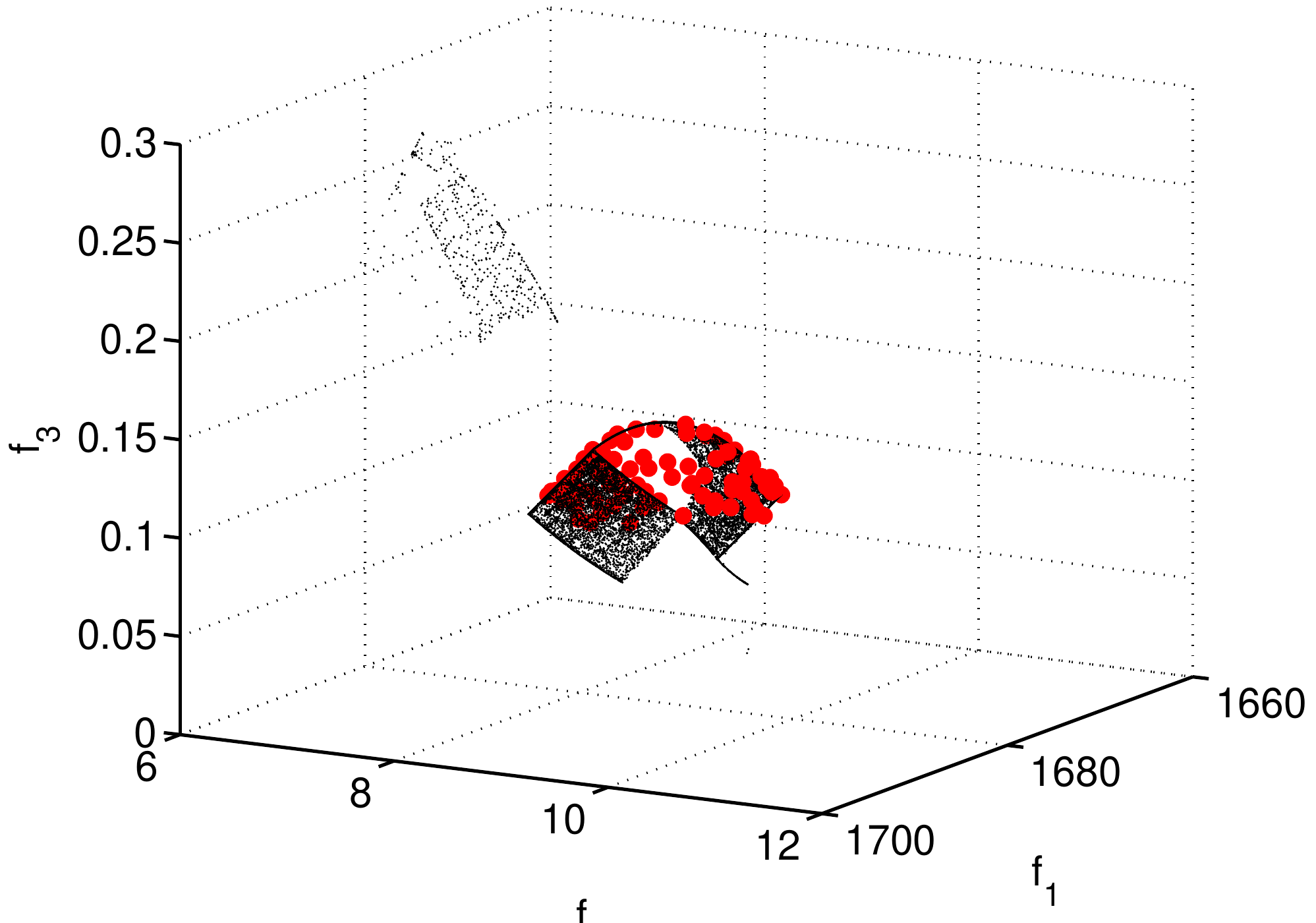}}
    \subfloat[NSGA-III]{
		\label{fig:cwdp.nsgaiii}
        \centering
		\includegraphics[width=1.7in]{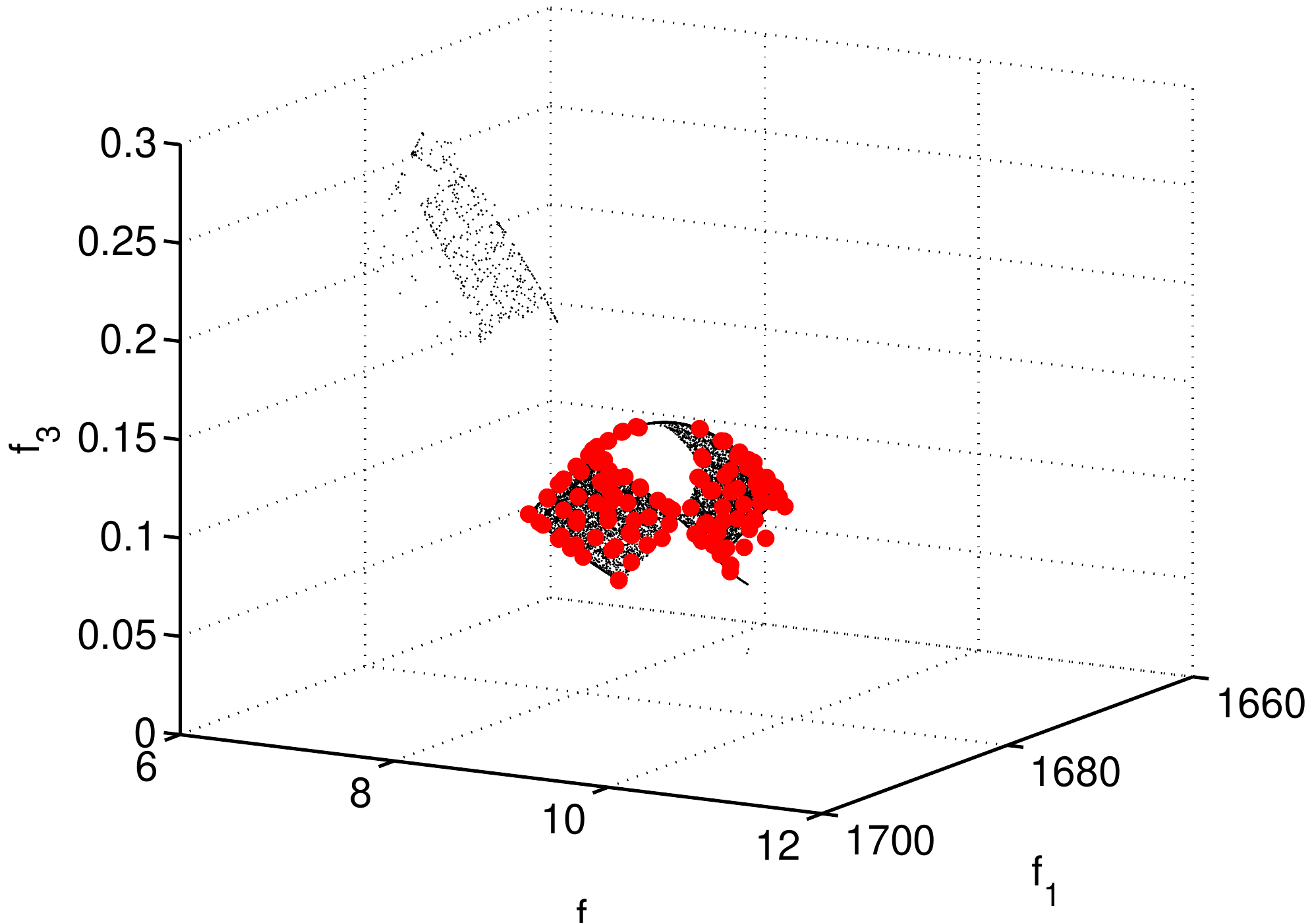}}
    \subfloat[MaOEA-CS]{
		\label{fig:cwdp.maoeacs}
        \centering
		\includegraphics[width=1.7in]{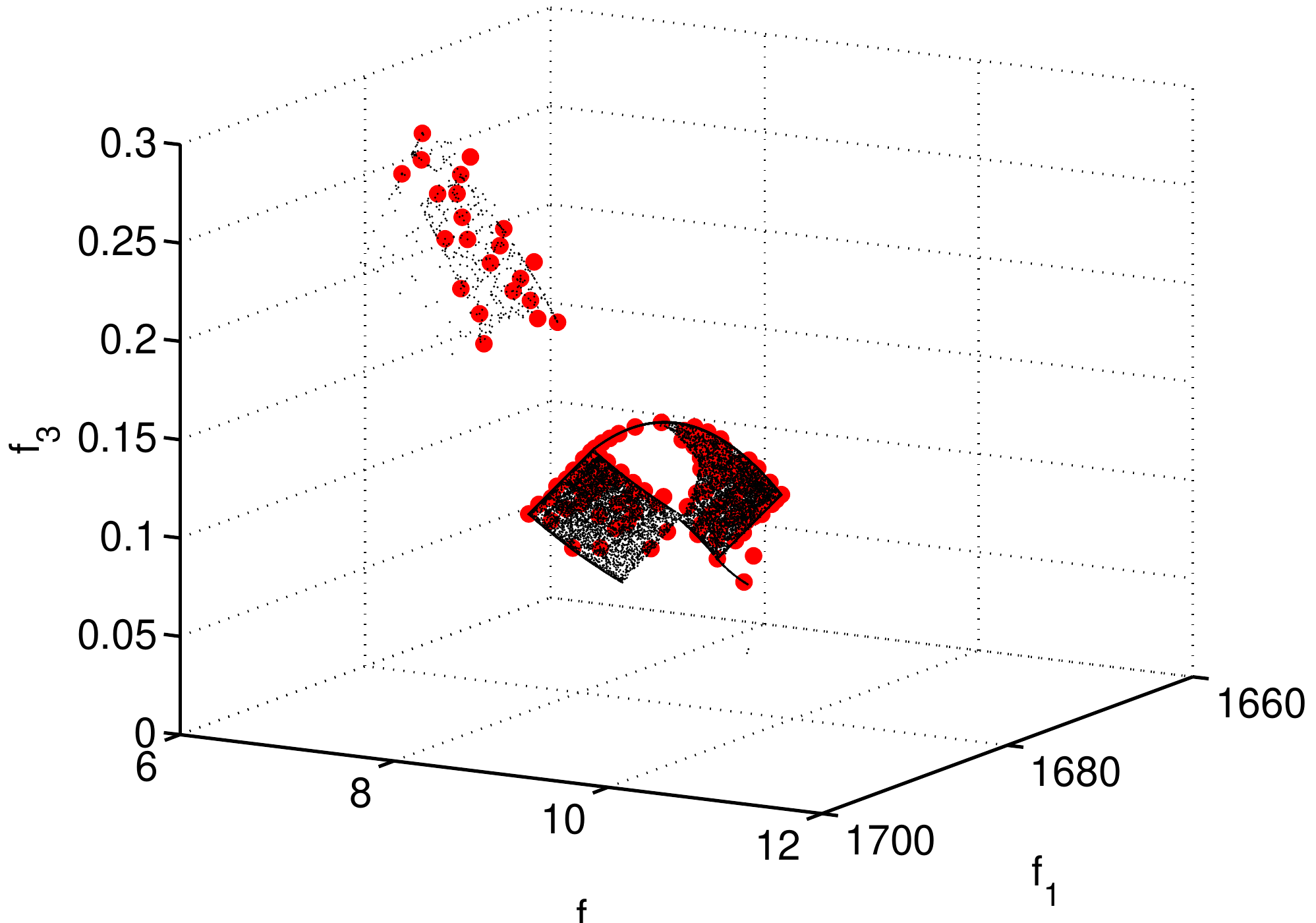}}
  \caption{The nondominated solutions obtained by all the seven compared algorithms on CWDV (marked by red circle), the small black dots represent PF approximations obtained by all the seven compared algorithms over all 30 runs.}\label{fig:CWDV}
\end{figure*}

The performance of all the seven compared algorithms, in terms of IGD and HV, are given in Table~\ref{tab:real-resluts}. For CWDV, MaOEA-CS obtains the best performance in terms of both IGD and HV. To visualize the performance of all the compared algorithms, the nondominated solutions obtained by all the seven algorithms in the run with the median IGD values are plotted in Fig. \ref{fig:CWDV}. It can be observed that CWDV has a discontinuous PF and only MaOEA-CS can obtain the nondominated solutions on all parts of PFs.

\begin{table*}
\caption{Experimental results of 7 algorithms on CWDV and CSIP}\label{tab:real-resluts}
\centering
\resizebox{7in}{!}{
    \begin{tabular}{ccccccccc}
    \toprule
    Problem & indicator & BCE-MOEAD & GSRA  & KnEA  & RSEA  & RVEA  & NSGA-III & MaOEA-CS \bigstrut\\
    \midrule
    \multirow{2}[2]{*}{CWDV} & IGD & 3.342e-02~(4.4e-04)$^-$ & 1.780e-01~(1.2e-04)$^-$ & 5.192e-02~(4.9e-03)$^-$ & 3.466e-02~(8.0e-04)$^-$ & 6.134e-02~(1.3e-02)$^-$ & 3.822e-02~(3.0e-03)$^-$ & \cellcolor[rgb]{0.749, 0.749, 0.749}1.831e-02~(6.8e-04) \\
          & HV    & 9.800e-01~(3.0e-03)$^-$ & 9.251e-01~(1.1e-03)$^-$ & 9.676e-01~(4.7e-03)$^-$ & 9.770e-01~(3.2e-03)$^-$ & 9.533e-01~(1.3e-02)$^-$ & 9.765e-01~(2.7e-03)$^-$ & \cellcolor[rgb]{0.749, 0.749, 0.749}1.017e+00~(1.0e-03)  \bigstrut[b]\\
    \midrule
    \multirow{2}[2]{*}{CSIP} & IGD    & \cellcolor[rgb]{0.749, 0.749, 0.749}1.761e-01~(3.4e-03)$^+$ & 3.105e-01~(1.5e-02)$^-$ & 1.909e-01~(1.6e-02)$^+$ & 3.008e-01~(5.0e-02)$^-$ & 3.444e-01~(1.9e-02)$^-$ & 1.955e-01~(6.7e-03)$^-$ & 1.937e-01~(1.0e-02) \\
          & HV    & 1.653e-01~(1.0e-02)$^-$ & 1.182e-01~(1.7e-02)$^-$ & 1.609e-01~(1.6e-02)$^-$ & 2.196e-01~(1.0e-02)$^-$ & 8.066e-02~(1.9e-02)$^-$ & 1.533e-01~(1.7e-02)$^-$ & \cellcolor[rgb]{0.749, 0.749, 0.749}2.569e-01~(1.0e-02)  \bigstrut[b]\\
    \bottomrule
    \end{tabular}
    }
\end{table*}

\begin{figure*}
  \centering
  	\subfloat[BCE-MOEA/D]{
		\label{fig:csip.bcemoead}
        \centering
		\includegraphics[width=1.7in]{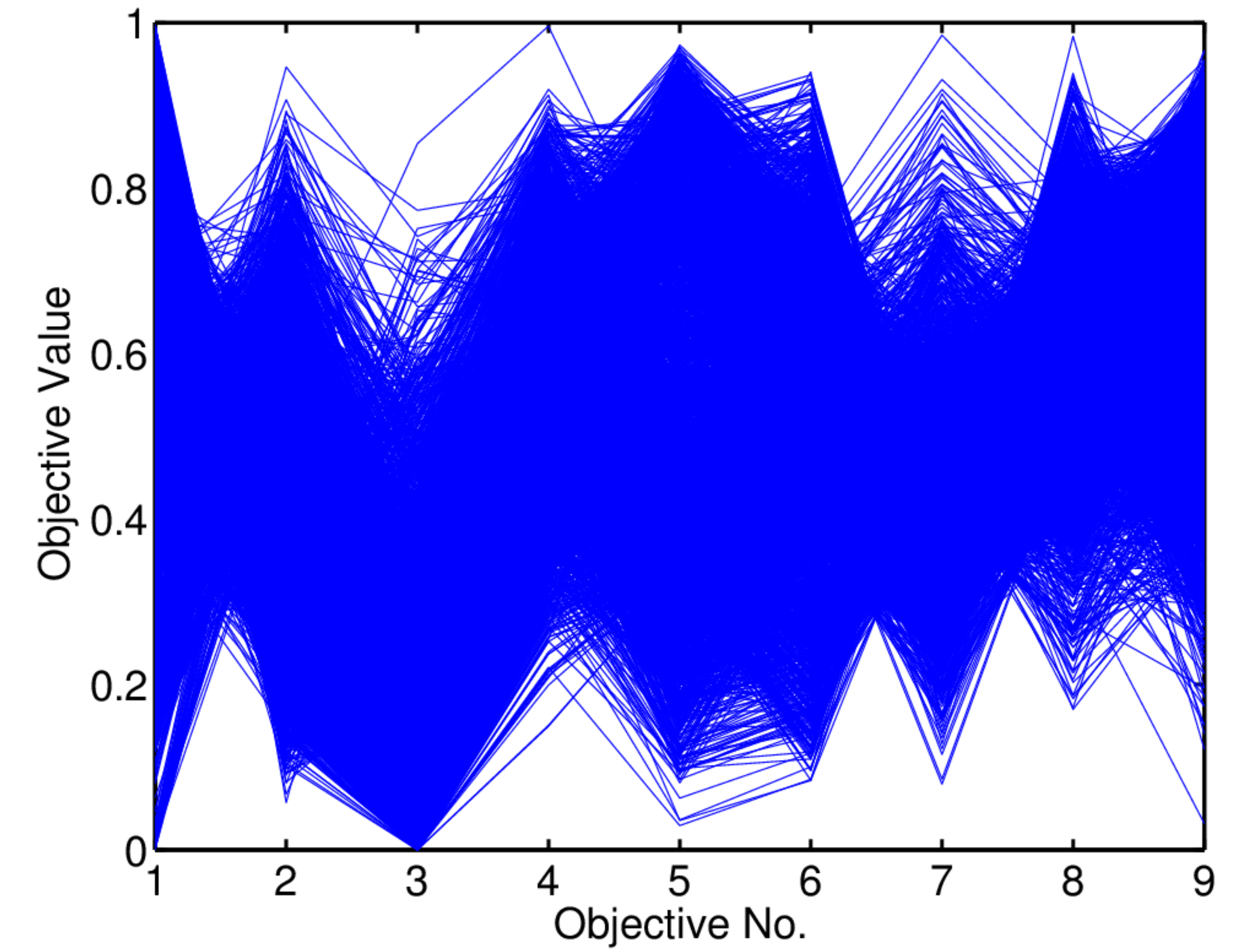}}
	\subfloat[GSRA]{
		\label{fig:csip.gsra}
        \centering
		\includegraphics[width=1.7in]{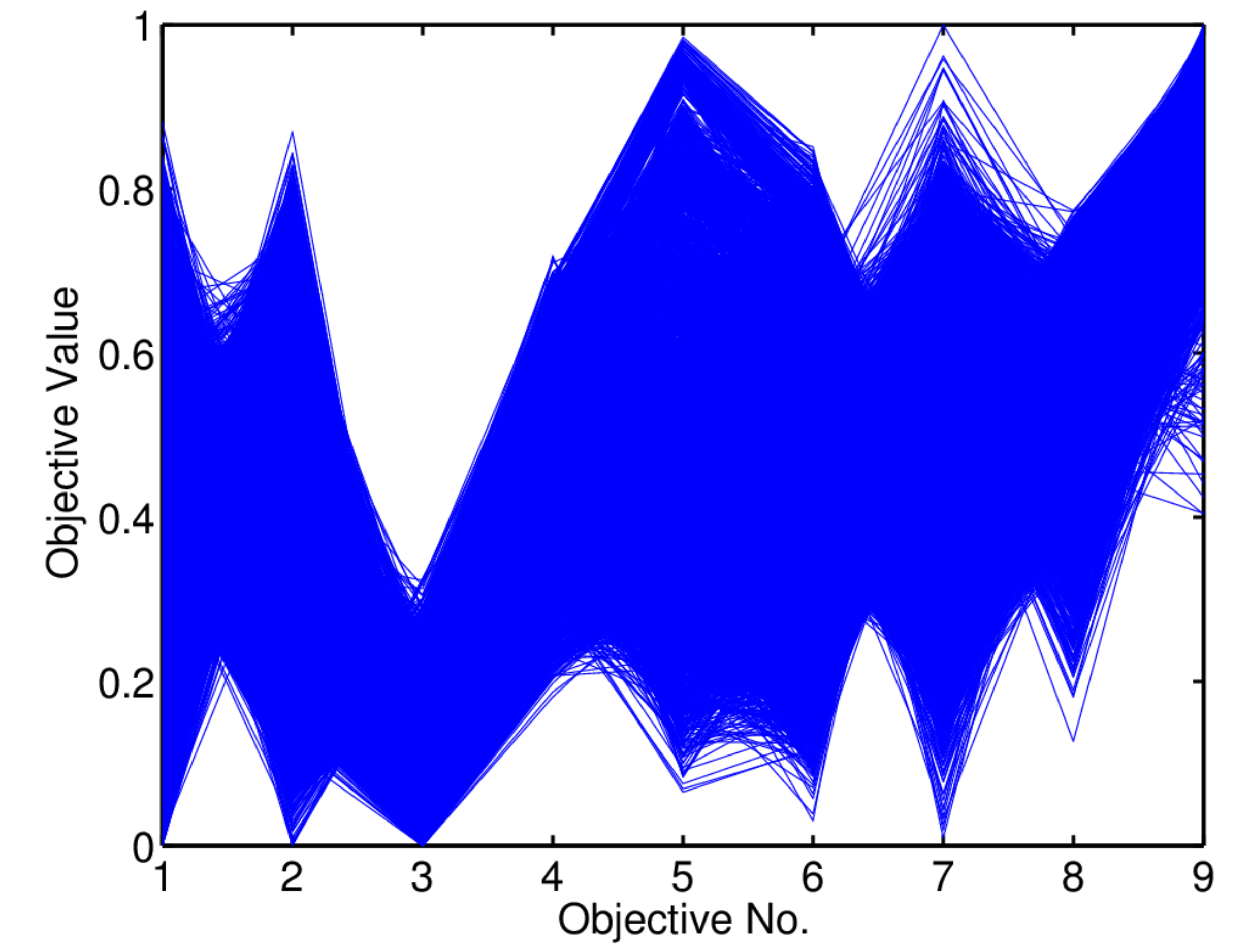}}
    \subfloat[KnEA]{
		\label{fig:csip.knea}
        \centering
		\includegraphics[width=1.7in]{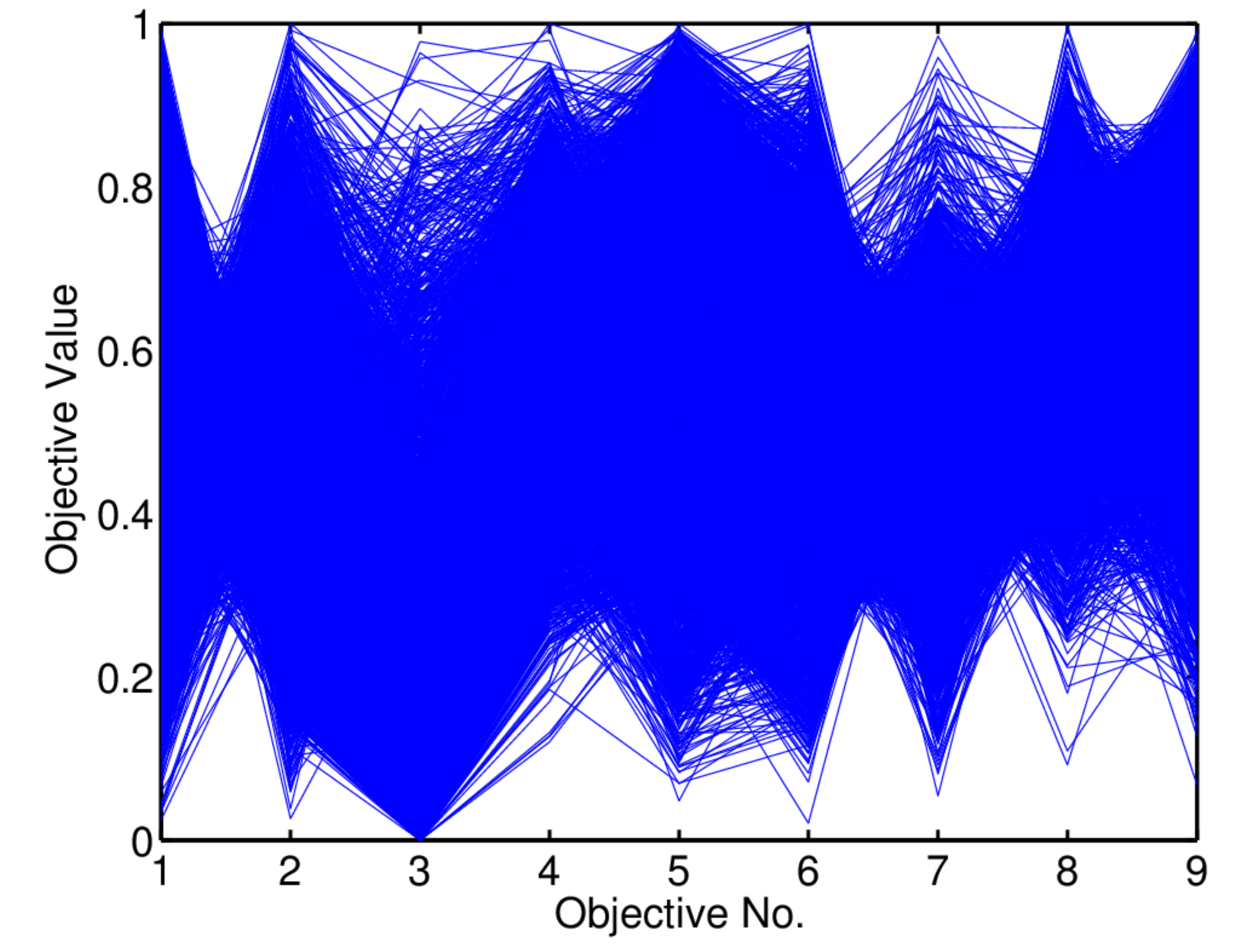}}
    \subfloat[RSEA]{
		\label{fig:csip.rsea}
        \centering
		\includegraphics[width=1.7in]{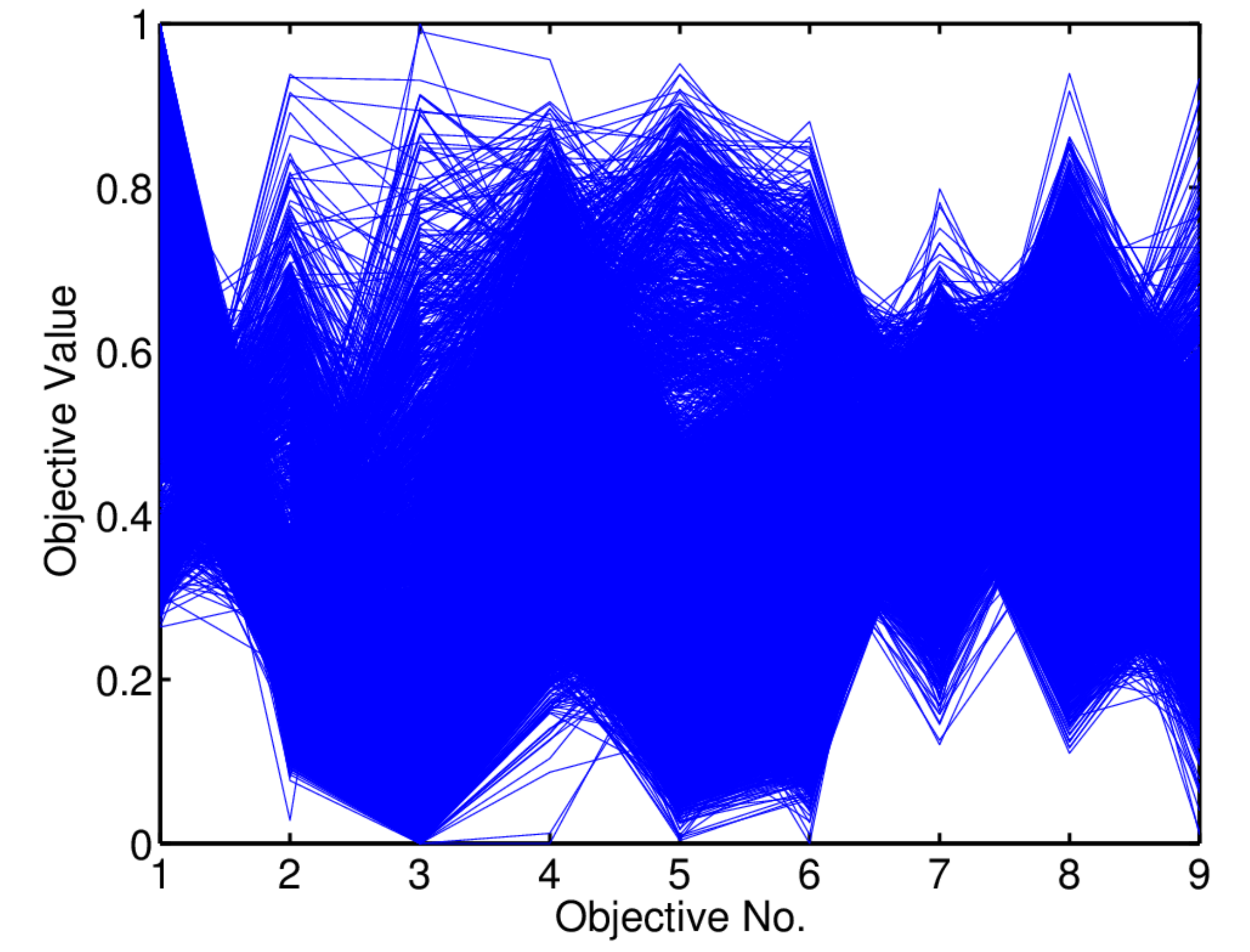}}

    \subfloat[RVEA]{
		\label{fig:csip.rvea}
        \centering
		\includegraphics[width=1.7in]{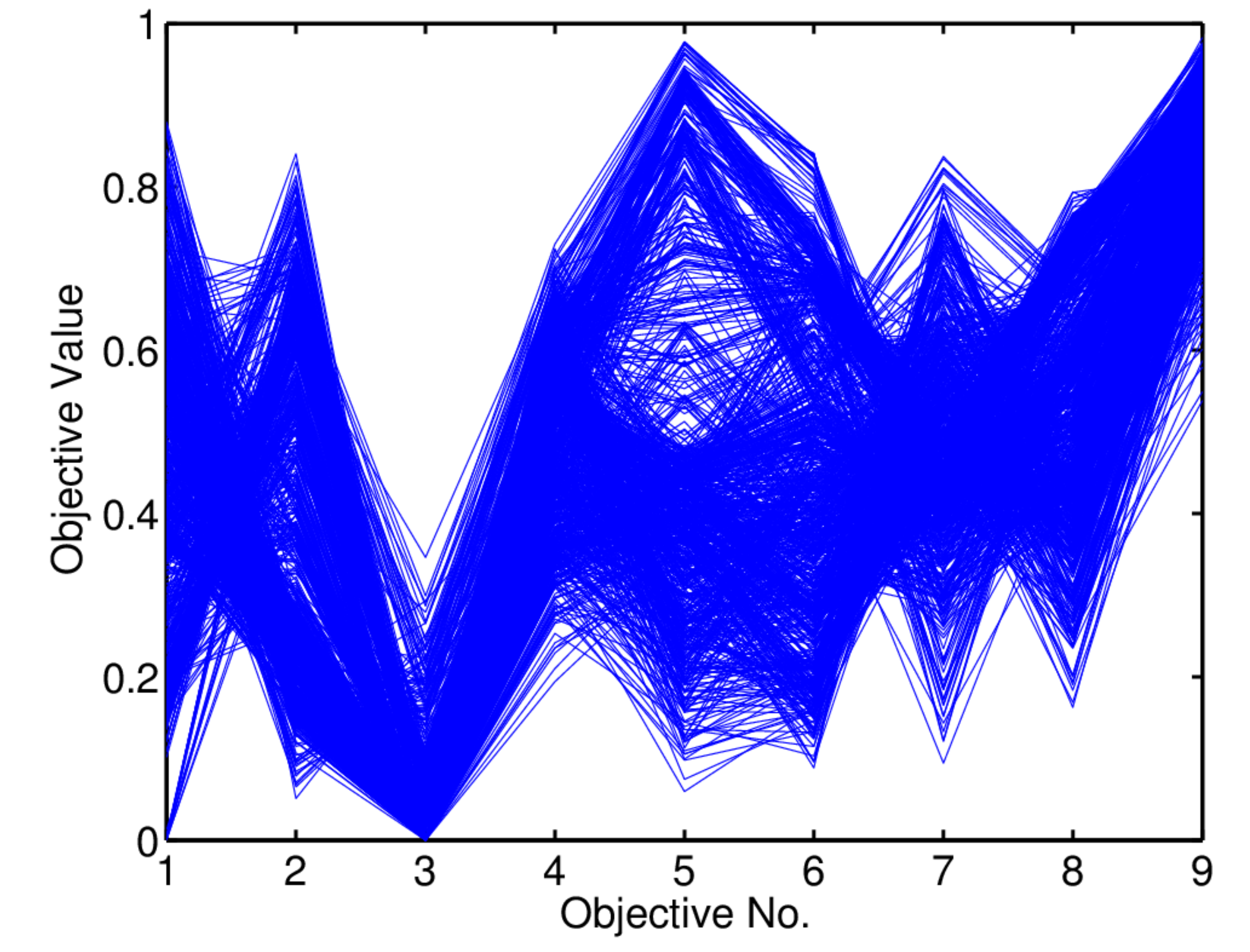}}
    \subfloat[NSGA-III]{
		\label{fig:csip.nsgaiii}
        \centering
		\includegraphics[width=1.7in]{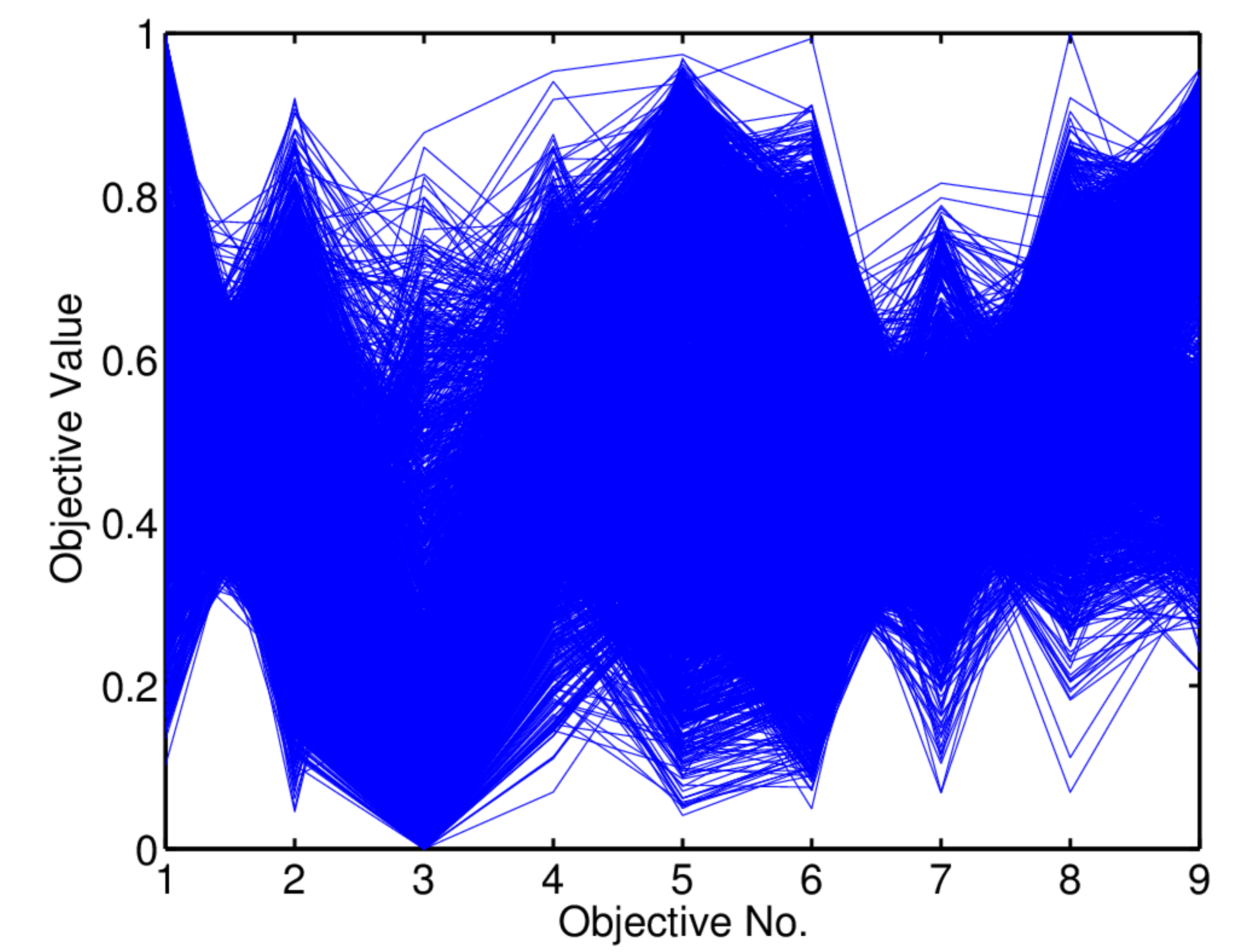}}
    \subfloat[MaOEA-CS]{
		\label{fig:csip.maoeacs}
        \centering
		\includegraphics[width=1.7in]{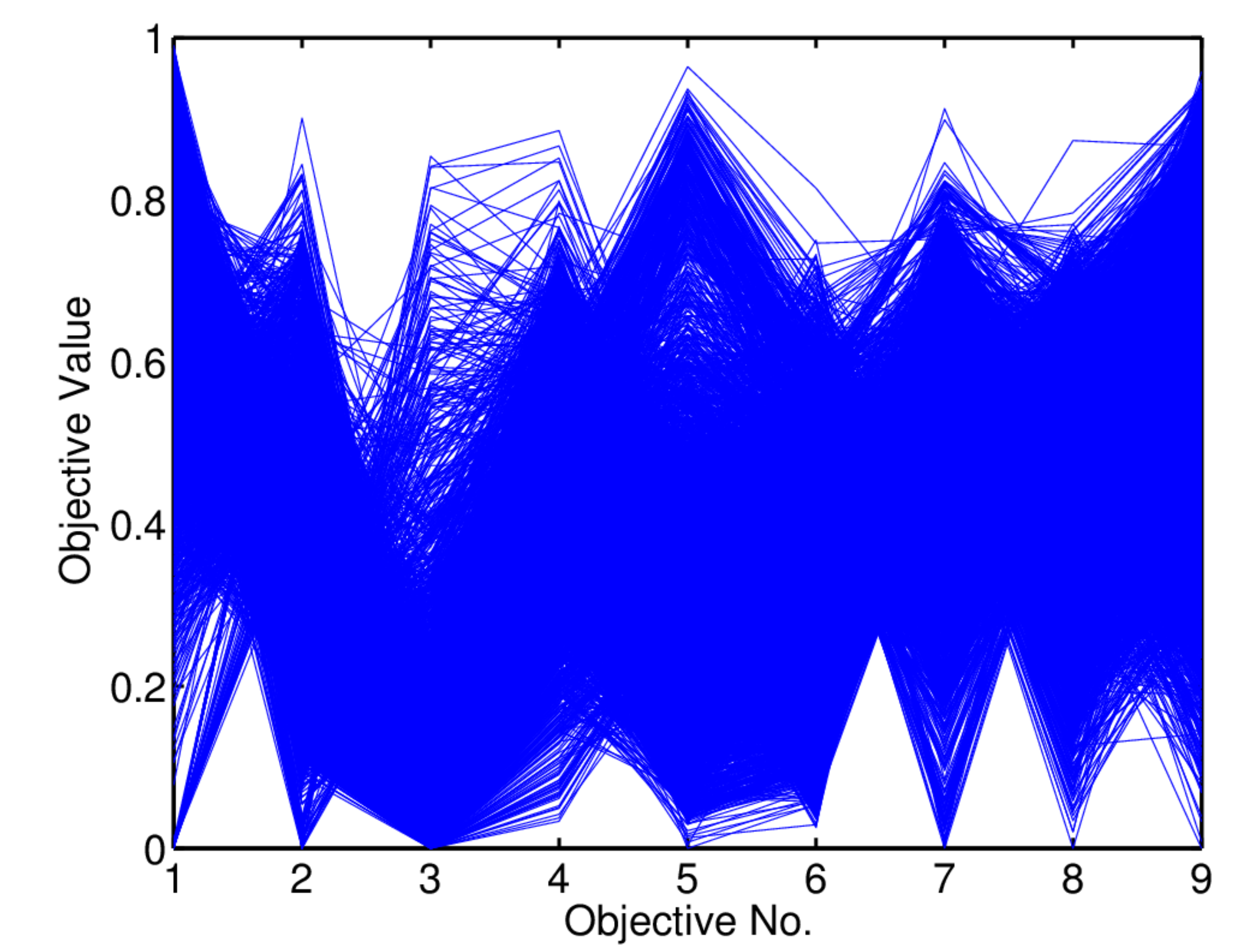}}
    \subfloat[Reference PF]{
		\label{fig:csip.pf}
        \centering
		\includegraphics[width=1.7in]{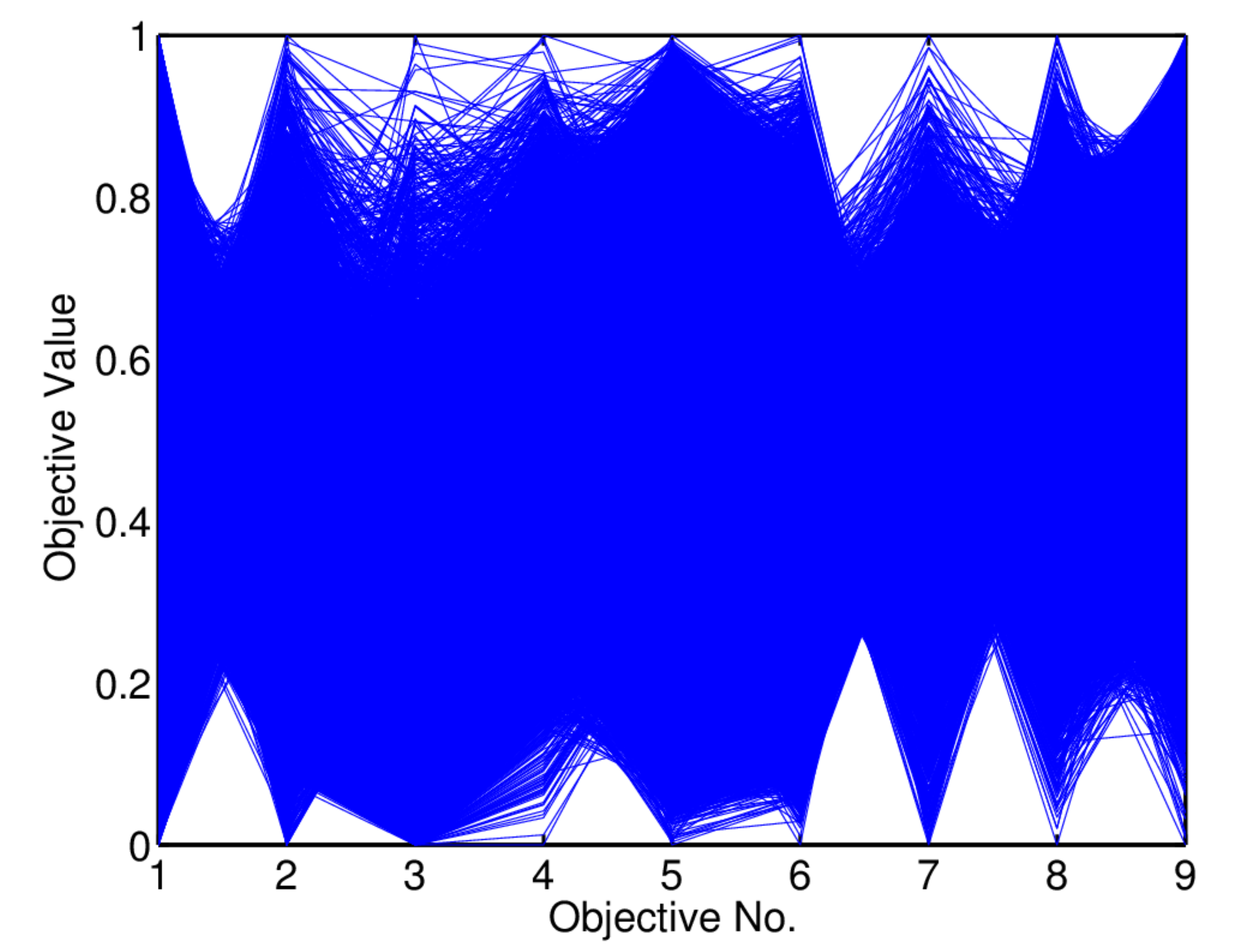}}
  \caption{The parallel coordinate plots of all the non-dominated solutions obtained by seven compared algorithms on CSIP over 30 runs. Among them, Fig. \ref{fig:csip.pf} presents the parallel coordinate plots of reference PF approximations obtained by all the seven compared algorithms over 30 runs.}\label{fig:CSIP}
\end{figure*}

As for CSIP, MaOEA-CS has the best performance in terms of HV while BCE-MOEAD has the best performance in terms of IGD. To further visualize the performance of all the compared algorithms, the parallel coordinate plots of all the nondominated solutions obtained by the seven algorithms over 30 runs are illustrated in Fig. \ref{fig:CSIP}. A reference PF is approximated  by using all the nondominated solutions obtained by all the seven algorithms over 30 runs, as shown in Fig. \ref{fig:csip.pf}.  It can be observed that only MaoEA-CS is able to obtain solutions with the minimum objective values on all the objectives. This can be explained by the fact that MaOEA-CS is able to locate the boundary of PFs by the approximation of the corner solutions. This also explains why MaOEA-CS can achieve better performance in terms of HV.

In addition, it also can be observed in Fig. \ref{fig:csip.bcemoead} and Fig. \ref{fig:knea} that BCE-MOEA/D and KnEA are able to obtain more diversely-populated solutions inside the PF boundaries. As IGD indicator may prefer solutions located inside the PF boundaries, BCE-MOEA/D and KnEA achieves better performance in terms of IGD on CSIP.

All the above experimental studies indicate that MaoEA-CS has satisfactory performance on the real-world optimization problems which usually have very irregular PFs.

\section{Conclusion}
In this paper, a many-objective evolutionary algorithm based on corner solution search was proposed and compared with six state-of-the-art algorithms on MaF test suite with various different characteristics for CEC$^\prime$2017 MaOEA Competition. The experimental results show that the proposed MaOEA-CS has the best overall performance. This indicates MaOEA-CS is more robust than other compared algorithms on various test problems thus it successfully won the CEC$^\prime$2017 MaOEA Competition.  The sensitivity test of two parameters in MaOEA-CS were also conducted and analyzed in this paper. In addition, MaOEA-CS has also been applied on two real-world engineering optimization problems with very irregular PFs. The experimental results show that MaOEA-CS outperforms other six compared algorithms in terms of either convergence or diversity, which indicates it has the ability to handle real-world complex optimization problems with irregular PFs.

\bibliographystyle{ieeetr}
\bibliography{combined}
\end{document}